\documentclass[lettersize,journal]{IEEEtran}
\usepackage{amsmath,amsfonts}
\usepackage{array}
\usepackage[caption=false,font=normalsize,labelfont=sf,textfont=sf]{subfig}
\usepackage{textcomp}
\usepackage{stfloats}
\usepackage{url}
\usepackage{verbatim}
\usepackage{graphicx}
\usepackage{booktabs}
\usepackage{bbding}
\usepackage{siunitx} 
\usepackage{amsmath}                                        
\usepackage[acronym]{glossaries}                            
\usepackage{algorithm}
\usepackage{multirow}    
\usepackage{amssymb} 
\usepackage{algpseudocode}
\usepackage{booktabs} 
\usepackage{graphicx} 
\usepackage{caption}  
\usepackage{colortbl}
\usepackage{wrapfig}
\usepackage{xcolor}
\definecolor{mygray}{gray}{0.6}
\usepackage{titletoc}
\definecolor{academicblue}{RGB}{21, 101, 192} 
\definecolor{academicred}{RGB}{178, 34, 34}  

\usepackage{hyperref}
\hypersetup{
	colorlinks=true,       
	linkcolor=academicred, 
	citecolor=academicblue,
	urlcolor=red,        
	pdfborder={0 0 0},     
	bookmarks=true,        
	bookmarksopen=true,    
	pdftitle={Generative Semantic Multi-Object Tracking: A Large-Scale Benchmark and Macro-Cognitive Framework}, 
	pdfauthor={Pan Liao}   
}
\providecommand{\authcount}[1]{}

\newcommand{\best}[1]{\textbf{\textcolor{red}{#1}}}
\newcommand{\second}[1]{\textbf{\textcolor{blue}{#1}}}

\usepackage[accsupp]{axessibility}  
\usepackage{enumitem}
\usepackage{tcolorbox}

\def\BibTeX{{\rm B\kern-.05em{\sc i\kern-.025em b}\kern-.08em
		T\kern-.1667em\lower.7ex\hbox{E}\kern-.125emX}}
\usepackage{balance}

\begin{document}
\title{Generative Semantic Multi-Object Tracking: A Large-Scale Benchmark and an MLLM-Driven Reasoning Framework}
	
\author{Pan Liao, Feng Yang, Jinwen Yu, Di Wu, Wang Zhao and Dingwen Zhang
	\thanks{This work was supported by the National Natural Science Foundation of China under Grant No. 62293543. \textit{(Corresponding author: Feng Yang.)}}%
	\thanks{F. Yang, D. Wu, J. Yu, and D. Zhang are with the School of Automation, Northwestern Polytechnical University, Xi'an, China (e-mail: yangfeng@nwpu.edu.cn).}%
	\thanks{P. Liao is with the School of Data Science, Lingnan University, Hong Kong (e-mail: liaopan@mail.nwpu.edu.cn).}%
	\thanks{W. Zhao is with the Department of Mechanical Engineering, Tsinghua University, Beijing, China (e-mail: 18992368054@mail.nwpu.edu.cn).}%
}

\markboth{IEEE TRANSACTIONS ON IMAGE PROCESSING,~VOL.~XX, NO.~XX, 2026}%
{Liao \MakeLowercase{\textit{et al.}}: Generative Semantic Multi-Object Tracking}
	
	
	\maketitle
	
\begin{abstract}
		Semantic Multi-Object Tracking (SMOT) is evolving from purely geometric localization toward comprehensive video understanding. However, existing paradigms predominantly rely on closed-set interaction tags and fragmented perception pipelines, creating a bottleneck that prevents the full utilization of Multi-modal Large Language Models (MLLMs) for dynamic scenes. In this paper, we elevate SMOT from rigid classification to an open-ended generative reasoning task. To support this paradigm shift, we introduce Grand-SMOT, a large-scale benchmark featuring high-density, dual-stream narratives. This dataset explicitly decouples micro-level individual dynamics from macro-level environmental contexts, directly resolving the semantic scarcity of prior tracking datasets. Furthermore, we propose LLMTrack, a unified MLLM-driven framework for dynamic SMOT. Guided by a verifiable ``\textit{Macro-Understanding-First}'' mechanism, LLMTrack employs a Spatio-Temporal Fusion Module to compress discrete geometric trajectories into continuous semantic tokens, effectively suppressing temporal hallucinations in long-sequence tracking. Extensive experiments, utilizing a novel decoupled evaluation protocol, validate that LLMTrack achieves state-of-the-art geometric tracking robustness while delivering a qualitative leap in generative semantic reasoning. The code and datasets are publicly available at \url{https://github.com/liaopan-lp/LLMTrack-GrandSMOT}.
\end{abstract}
	

	\section{Introduction}

	\IEEEPARstart{M}{ulti}-Object Tracking (MOT) is evolving from geometric localization to Semantic MOT (SMOT) \cite{li2024beyond}. By answering not only ``\textit{where objects are}'' but also ``\textit{what they are doing}'' and ``\textit{what happens contextually},'' SMOT aligns with the broader goals of video analytics. Formulating SMOT as a generative reasoning task bridges the gap between passive visual perception \cite{li2024saliency,yang2025geodesic} and continuous scene understanding \cite{gao2026cosurfgs, fang2025associate,ling2026identity}. This transition is essential for developing intelligent agents in applications such as Embodied AI and autonomous driving, which require active, causal interpretation of the dynamic physical world \cite{bi2026revisiting}.
	
	\begin{figure}[t]
		\centering
		\includegraphics[width=1\linewidth]{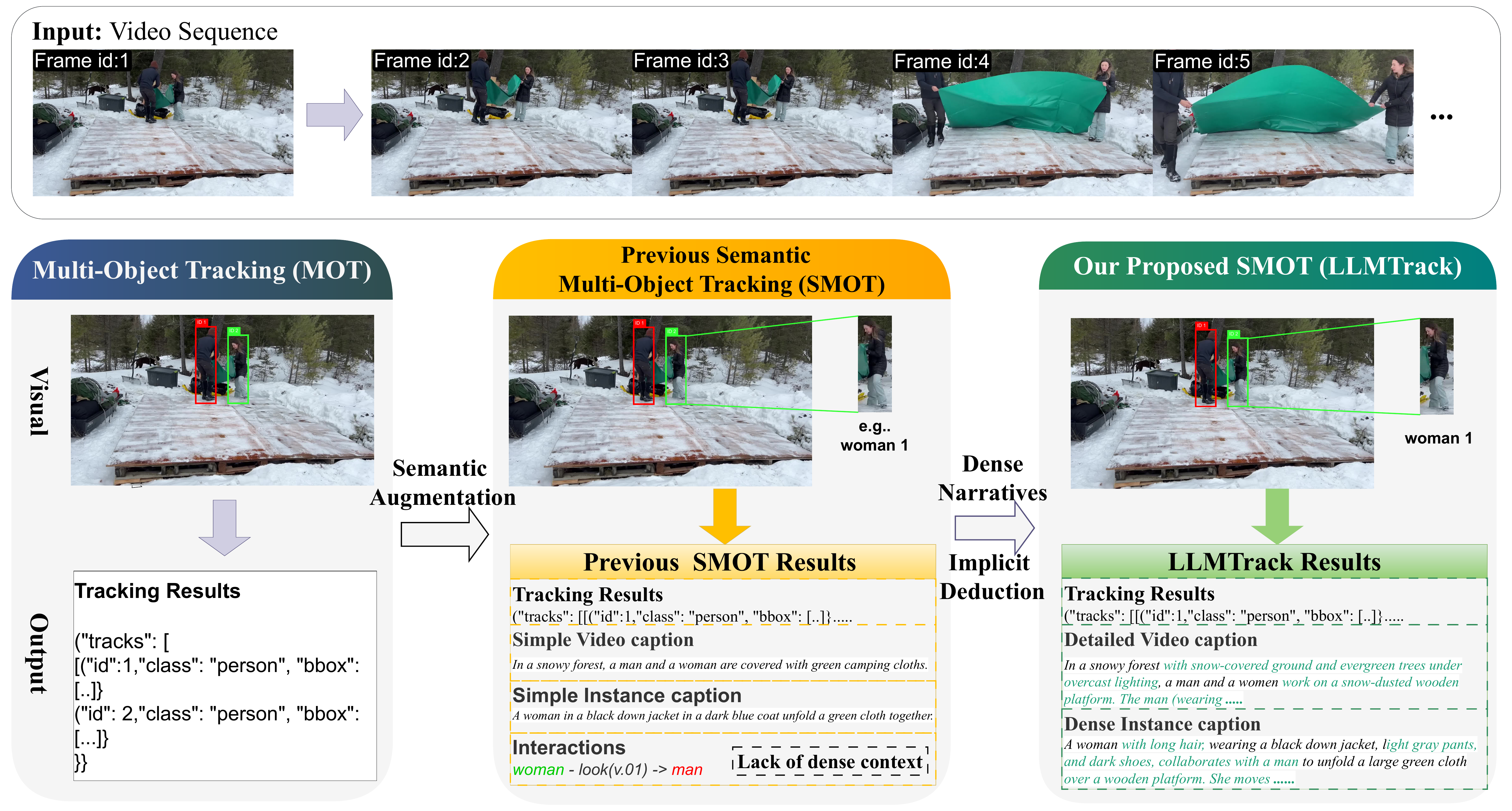} 
		\caption{Conceptual comparison among MOT, previous SMOT, and our LLMTrack. Previous SMOT treats interactions as isolated closed-set tags with simple captions, lacking dense context. LLMTrack instead unifies tracking, environment, and interactions into dense narratives via implicit logical deduction.}
		\label{fig:teaser}
	\end{figure}
	
	However, current SMOT paradigms exhibit structural limitations across task formulation, data infrastructure, and architectural design. To address these issues, this paper investigates how SMOT can transition from closed-set interaction classification to open-ended semantic reasoning. Concurrently, we examine the data infrastructure required to benchmark context-behavior decoupling, and how Multi-modal Large Language Models (MLLMs)~\cite{yang2025qwen3,li2024llava,team2025minicpm4,wang2025internvl3} can be effectively adapted for continuous dynamic tracking.
	
	First, existing methodologies \cite{li2024beyond} typically treat object interactions as predefined, closed-set classification tasks, leaving tracking pipelines semantically constrained. Rather than treating interaction as independent metadata to be memorized via label injection, we argue that it is a logical deduction emerging naturally from the interplay between individual behaviors and environmental contexts. As illustrated in \textbf{Fig.~\ref{fig:teaser}}, we propose shifting the SMOT task from discriminative tag prediction to generative narrative reasoning. To evaluate this capability, we introduce a decoupled evaluation protocol designed to strictly assess the holistic semantic fidelity and reasoning logic of generated narratives, isolating them from geometric tracking penalties.
	
	This shift in task formulation requires a corresponding update in data infrastructure. Existing SMOT benchmarks rely on shallow tags or single-sentence descriptions that fail to capture the video-level atmosphere or fine-grained dynamics \cite{chen2024sharegpt4v}. To address this semantic scarcity, we introduce Grand-SMOT, a large-scale, open-world benchmark. By integrating the real-world complexity of TAO \cite{dave2020tao} with the semantic foundation of BenSMOT \cite{li2024beyond}, we implement a dual-stream dense annotation strategy. This approach explicitly decouples individual dynamics from the environmental atmosphere, establishing new evaluation metrics for long-term causal reasoning and context-behavior consistency.
	
	Finally, integrating MLLMs into continuous tracking poses architectural challenges, particularly regarding temporal consistency. To address this, we propose LLMTrack, an MLLM-driven framework for generative semantic tracking. LLMTrack utilizes Grounding DINO \cite{liu2024grounding} for visual anchoring and introduces a Spatio-Temporal Fusion Module that maps discrete geometric trajectories into continuous semantic tokens. Crucially, the generation process is guided by a verifiable ``\textit{Macro-Understanding-First}'' mechanism, which structures the MLLM prompt to anchor the global environmental context strictly before evaluating micro-level instance dynamics. This causal token ordering adapts the MLLM for dynamic tracking by explicitly mitigating temporal hallucinations and spurious interactions.
	
	In summary, our contributions are threefold:
	
\begin{enumerate}
	\item \textbf{Core Paradigm: Emergent Interaction Reasoning.} We elevate Semantic MOT to an open-ended generative reasoning task. By introducing a ``\textit{Macro-Understanding-First}'' mechanism, we enable multi-object interactions to emerge naturally as logical deductions, effectively suppressing temporal hallucinations.
	
	\item \textbf{Evaluation Platform: The Grand-SMOT Benchmark.} To validate this paradigm shift, we construct Grand-SMOT. Featuring dual-stream dense narratives and a novel decoupled evaluation protocol, it rigorously isolates and benchmarks semantic reasoning apart from geometric tracking errors.
	
	\item \textbf{Enabling Mechanism: Spatio-Temporal Fusion.} To deploy this paradigm in continuous video streams, we propose a Spatio-Temporal Fusion Module. By compressing discrete visual trajectories into continuous semantic tokens, it effectively bridges high-frequency visual perception with the cognitive capacity of MLLMs.
\end{enumerate}


	\begin{figure}[htbp]
		\centering
		\includegraphics[width=\linewidth]{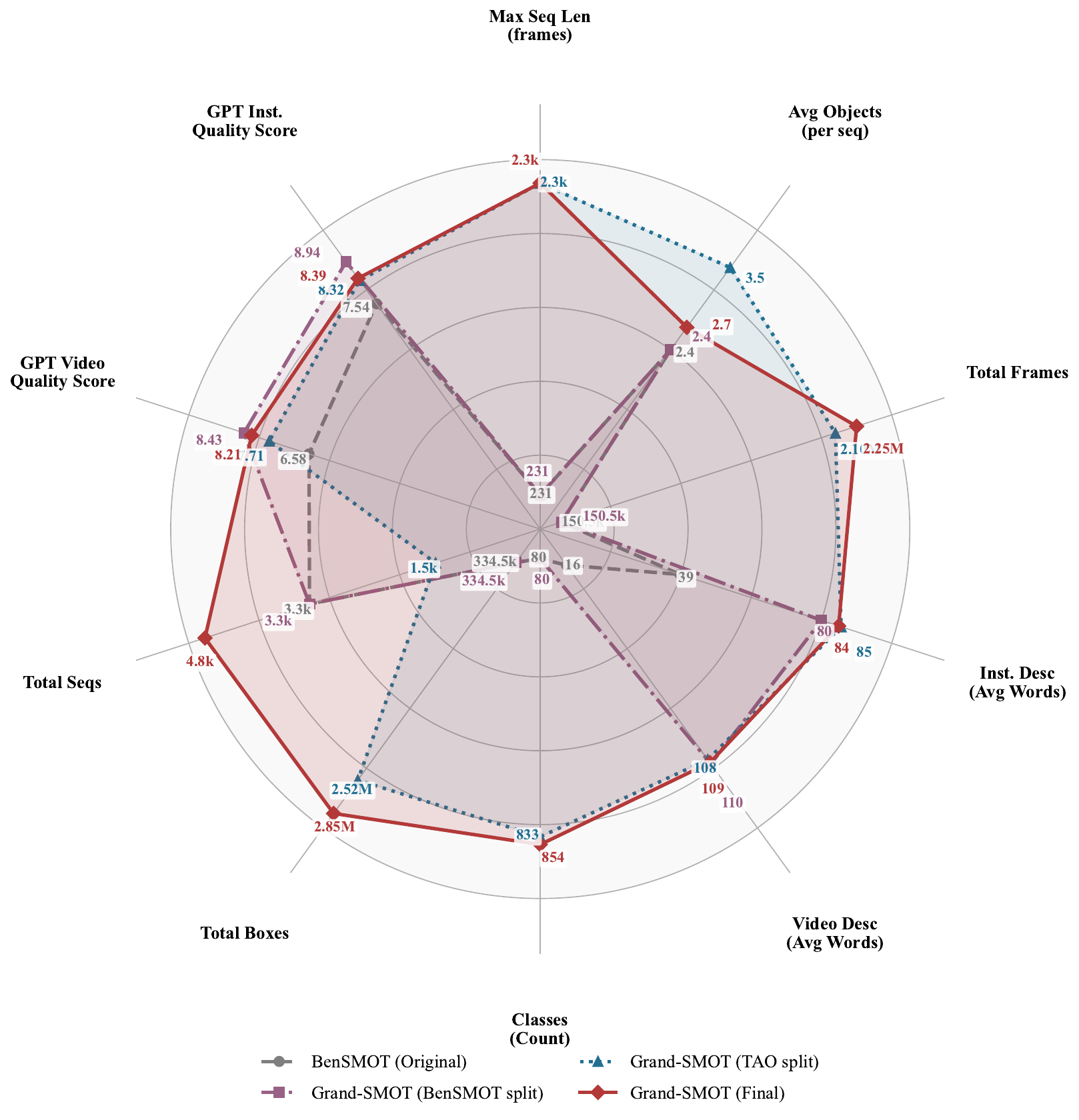}
		\vspace{-5pt}
		\caption{\textbf{Statistical Comparison of the Grand-SMOT Benchmark.} Radar chart comparing data distribution against BenSMOT \cite{li2024beyond}. Grand-SMOT achieves a superior balance in sequence length, instance density, and semantic richness.}
		\label{fig:dataset_radar}
		\vspace{-0.2cm}
	\end{figure}
	\section{Related Work}
	
	\subsection{From Discriminative Classification to Generative Reasoning}
	Traditional MOT has predominantly relied on geometric heuristics \cite{bewleySimpleOnlineRealtime2016, wojke2017simple} and fully learned association models \cite{zhangByteTrackMultiobjectTracking2022, liao2025fasttracktr, zeng2022motr, zhang2023motrv2}. While methods like ByteTrack \cite{zhangByteTrackMultiobjectTracking2022} achieve remarkable identity preservation, they remain semantically blind, representing objects merely as coordinate boxes. Recent paradigms have transitioned towards tracking-by-language methods, utilizing MLLMs for referring MOT \cite{wang2024language, lv2025vision} or leveraging adaptive state prompting \cite{zhang2025epiptrack}. Most notably, SMOTer \cite{li2024beyond} formalized Semantic MOT to simultaneously generate trajectories and captions. While SMOTer serves as a strong baseline for capturing object states, its reliance on discriminative models \cite{zhai2023siglip} intrinsically limits interaction recognition to a closed-set classification problem. This discriminative constraint hinders the open-world generative reasoning necessary for unconstrained, dynamic video understanding. Our work shifts this paradigm, proposing that interactions should be generated dynamically via an MLLM rather than classified from predefined tags.
	
	\subsection{Video-Level MLLMs vs. Track-Level Dynamics}
	MLLMs \cite{team2024gemini, an2025llava} have demonstrated profound zero-shot reasoning in static visual tasks \cite{liu2024improved, chen2024sharegpt4v}. However, directly applying frame-centric MLLMs \cite{fu2025llmdet} to continuous video streams invariably triggers temporal hallucinations and identity fragmentation. Conversely, video-native MLLMs \cite{li2025videochat} typically employ aggressive spatial-temporal token compression to manage computational costs. This compression irreversibly discards the fine-grained, track-level motion semantics essential for robust MOT. This architectural dilemma is further exacerbated by a critical data bottleneck: prevailing datasets \cite{dave2020tao, gupta2019lvis} supply only sparse categorical tags. Consequently, existing models either understand the global video but lose the object, or track the object but lack cognitive reasoning. To bridge this gap, our work synergizes the high-fidelity visual representations of LLaVA-OneVision \cite{li2024llava} with our Spatio-Temporal Fusion Module, which is specifically designed to maintain continuous object identities while feeding temporally coherent, uncompressed trajectory features into the MLLM.

	\section{The Grand-SMOT Dataset}
	\label{sec:dataset}

	\begin{figure*}[htbp]
		\centering
		\begin{minipage}{0.49\linewidth}
			\centering
			\includegraphics[width=\linewidth]{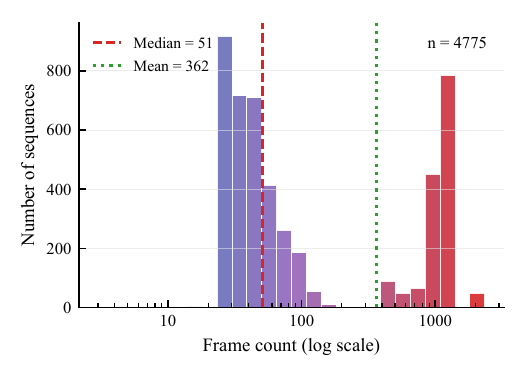}
			\vspace{2pt}
			\centerline{\small (a) Sequence Length Distribution}
		\end{minipage}
		\hfill
		\begin{minipage}{0.49\linewidth}
			\centering
			\includegraphics[width=\linewidth]{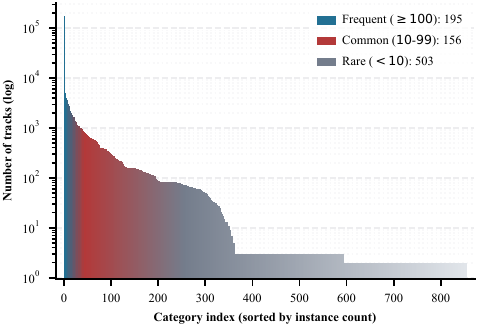}
			\vspace{2pt}
			\centerline{\small (b) Category Long-tail Distribution}
		\end{minipage}
		\vspace{-5pt}
		\caption{\textbf{Distributions of the Grand-SMOT Benchmark.} (a) The distribution of sequence lengths (frame counts) on a logarithmic scale. (b) The long-tail distribution of the tracking categories sorted by instance count.}
		\label{fig:dataset_histograms}
		\vspace{-0.2cm}
	\end{figure*}

Constructing a cognitive tracking benchmark requires transcending the geometric constraints of traditional datasets. We introduce \textbf{Grand-SMOT}, a large-scale semantic tracking benchmark designed to address the semantic scarcity in current MOT research. As summarized in Table~\ref{tab:dataset_comparison} and Figure~\ref{fig:dataset_radar}, Grand-SMOT distinguishes itself by providing high-density spatio-temporal narratives that support the proposed \textit{Macro-Understanding-First} paradigm, significantly outperforming existing benchmarks in both semantic granularity and scale. Crucially, the fine-grained statistical properties of our dataset are characterized in Figure~\ref{fig:dataset_histograms}, which explicitly details the multi-scale sequence length distribution alongside the inherent long-tail distribution across diverse tracking categories.
	
	\begin{figure*}[htbp]
		\centering
		\includegraphics[width=1\textwidth]{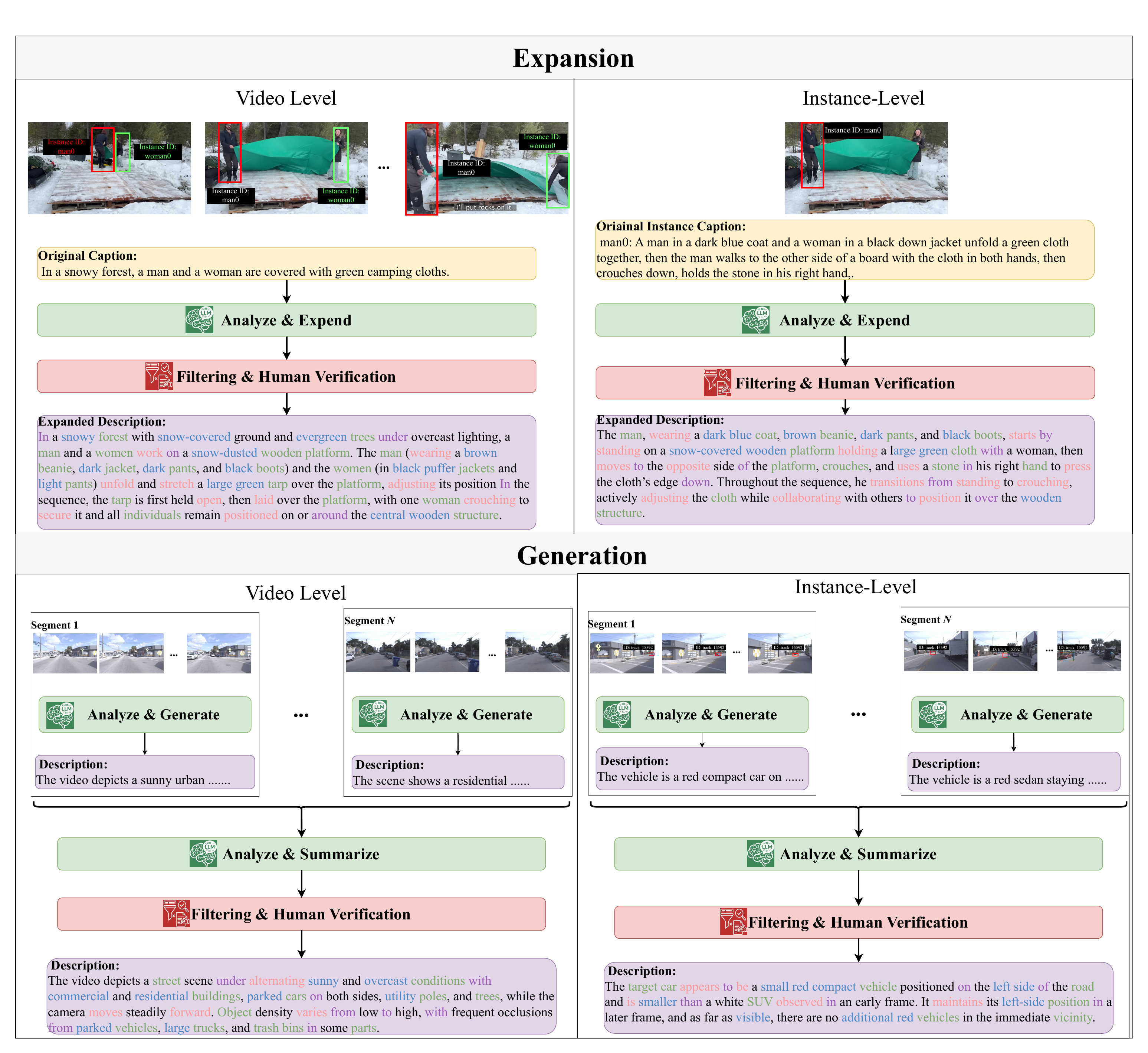}
		\caption{Our pipeline uses \textbf{Expansion} (top) to upgrade sparse labels to rich descriptions, and \textbf{Generation} (bottom) to synthesize video and instance dynamics via a segment-to-global approach.}
		\label{fig:generate_dataset}
	\end{figure*}
	
	\subsection{Data Formulation and Philosophy}
	Departing from prior methodologies that reduce interaction to rigid classification labels, we reframe interaction as a logical deduction emerging from the interplay between an agent and its environment. Formally, each training sample is defined as a triplet $\mathcal{S} = (V, \mathcal{T}_{env}, \mathcal{T}_{ins})$. Here, $V \in \mathbb{R}^{T \times H \times W \times 3}$ denotes the video sequence with $T$ frames, while $\mathcal{T}_{env}$ represents the \textbf{Video-Level Caption}, capturing the global atmosphere, weather conditions, and scene context. We explicitly decouple this from $\mathcal{T}_{ins} = \{ (b_i, d_i) \}_{i=1}^K$, the \textbf{Instance-Level Caption}, where $b_i \in \mathbb{R}^{L_i \times 4}$ represents the continuous bounding box sequence (with length $L_i$) of the $i$-th target, while $d_i$ chronicles its appearance evolution and fine-grained micro-actions. This dual-stream structure compels the tracker to implicitly reason about relations, rather than merely memorizing biased label co-occurrences.
	
	\subsection{Unified Generation Pipeline}
	Creating such dense annotations from scratch is prohibitively expensive. Therefore, we designed a unified pipeline (Figure~\ref{fig:generate_dataset}) that adapts to the distinct characteristics of two mainstream datasets: BenSMOT~\cite{li2024beyond} and TAO~\cite{dave2020tao}. We utilize Qwen3-VL-32B~\cite{yang2025qwen3}, currently one of the leading open-source MLLMs, as our core semantic engine to ensure high-quality text generation.

	\begin{table}[t]
		\centering
		\caption{\textbf{Comparison with representative MOT and tracking benchmarks.} \textbf{Inst. Desc.}: Instance-level semantic descriptions. \textbf{Env. Context}: Video-level environmental context. \textbf{Grand-SMOT} (Ours) stands out as the only benchmark providing dual-stream dense narratives with extensive open-world vocabulary.}
		\label{tab:dataset_comparison}
		\resizebox{\linewidth}{!}{%
			\begin{tabular}{l|c|ccc|cc}
				\toprule
				\multirow{2}{*}{\textbf{Dataset}} & \multirow{2}{*}{\textbf{Classes}} & \multicolumn{3}{c|}{\textbf{Scale}} & \multicolumn{2}{c}{\textbf{Semantic Granularity}} \\
				& & \textbf{Videos} & \textbf{Tracks} & \textbf{Frames} & \textbf{Inst. Desc.} & \textbf{Env. Context} \\
				\midrule 
				MOT17 \cite{milanMOT16BenchmarkMultiObject2016} & 1 & 14 & 1.3K & 11K & $\times$ & $\times$ \\
				MOT20 \cite{dendorfer2020mot20} & 1 & 8 & 3.8K & 13K & $\times$ & $\times$ \\
				DanceTrack \cite{sunDanceTrackMultiObjectTracking2022} & 1 & 100 & 990 & 106K & $\times$ & $\times$ \\
				GroOT \cite{nguyen2023type} & 8 & 1.5k & 13.3K & 2.2M & Brief Sent. & $\times$ \\
				TAO \cite{dave2020tao} & 833 & 1.5K & 17.3K & 2.1M & Class Name & $\times$ \\
				BenSMOT \cite{li2024beyond} & 80 & 3.3K & 7.8K & 151K & Interaction Tag & Brief Sent. \\
				\midrule
				\rowcolor{green!10} \textbf{Grand-SMOT (Ours)} & \textbf{854}$^\dagger$ & \textbf{4.8K}$^*$ & \textbf{25}K$^*$ & \textbf{2.3M}$^*$ & \textbf{Dense Narrative} & \textbf{Dense Narrative} \\
				\bottomrule
			\end{tabular}
		}
		\begin{flushleft}
			\scriptsize
			$^\dagger$: The number of non-overlapping classes in the merged TAO and BenSMOT taxonomy. \quad $^*$: Approximate sum of integrated sources (TAO + BenSMOT).
		\end{flushleft}
		\vspace{-0.2cm}
	\end{table}
	
	\noindent\textbf{(1) Semantic Expansion:} 
	We formulate the expansion of BenSMOT as a conditional generative mapping $\Phi: (s, v) \rightarrow \mathcal{N}_{dense}$, where a sparse interaction tag $s$ serves as a semantic anchor rooted in the visual context $v$. This mapping function $\Phi$ executes a \textbf{Semantic Densification} process: it conditions the narrative generation on extracted environmental priors $\mathcal{E}$ (\textit{e.g.}, atmospheric conditions, scene layout) and fine-grained visual attributes. By embedding the rigid interaction label into this continuous, high-dimensional context, $\Phi$ transforms isolated tags into cohesive, visually grounded narratives that vividly encapsulate the target's presence within the scene.
	
	\noindent\textbf{(2) Hierarchical Generation:} 
	For TAO dataset, we propose a bottom-up paradigm to resolve long-sequence forgetting. We discretize the video stream into non-overlapping segments $\mathbf{S} = \{S_k\}_{k=1}^N$. To rigorously prevent label leakage from TAO's noisy taxonomy (\textit{e.g.}, generic "\textit{baby}" labels), we define a visual-only generation function $d_k = \mathcal{G}(I(S_k, b) \mid \emptyset_{text})$, which explicitly conditions strictly on the visual region $I(\cdot)$ while enforcing an empty textual prior $\emptyset_{text}$. Subsequently, a global fusion operator $\mathcal{T}_{global} = \mathcal{F}(\{d_k\}_{k=1}^N)$ aggregates these fragmented local descriptions, synthesizing a coherent narrative that preserves instance identity and temporal continuity across the entire sequence.

	\subsection{Quality Assurance}
	\label{subsec:qa}
	Cognizant of the hallucination risks inherent in MLLMs and the prohibitive costs of full manual verification, we enforce a cost-effective quality assurance protocol. To mitigate self-evaluation bias, we decouple generation from verification by employing an independent Vision-Language Critic (\textit{e.g.}, MiniCPM-V 4.0~\cite{team2025minicpm4}). The Critic computes a cross-modal alignment score (1--10) to assess semantic fidelity, temporal coherence, and physical plausibility between each raw video and its generated narrative. Sequences scoring below an empirically selected threshold ($\tau=7.0$) are flagged as ``\textit{hard cases}'' for manual inspection. This filtering identified 762 videos. Expert annotators then spent approximately 70 hours inspecting these cases and correcting errors such as identity switches and phantom actions. This targeted Human-in-the-Loop process concentrates manual effort on uncertain samples while retaining exactly \textbf{4,775} verified sequences in the final benchmark.
	
	\subsection{Decoupled Evaluation Protocol}
	\label{subsec:decoupled_evaluation}
	To rigorously evaluate the model's pure semantic understanding capabilities on Grand-SMOT without unfairly compounding penalties from geometric tracking errors, we designed a \textbf{Decoupled Evaluation Strategy}. Standard evaluation protocols often apply a double penalty when tracking fails: the model is penalized once in the tracking metrics (\textit{e.g}., HOTA, IDF1) and again in the captioning metrics because the corresponding text is either missing or fragmented. To mitigate this interpretation penalty, we implemented two specific masking operations during metric calculation.
	
\begin{figure}[htbp]
		\vspace{-0.1cm}
		\centering
		\includegraphics[width=1\linewidth]{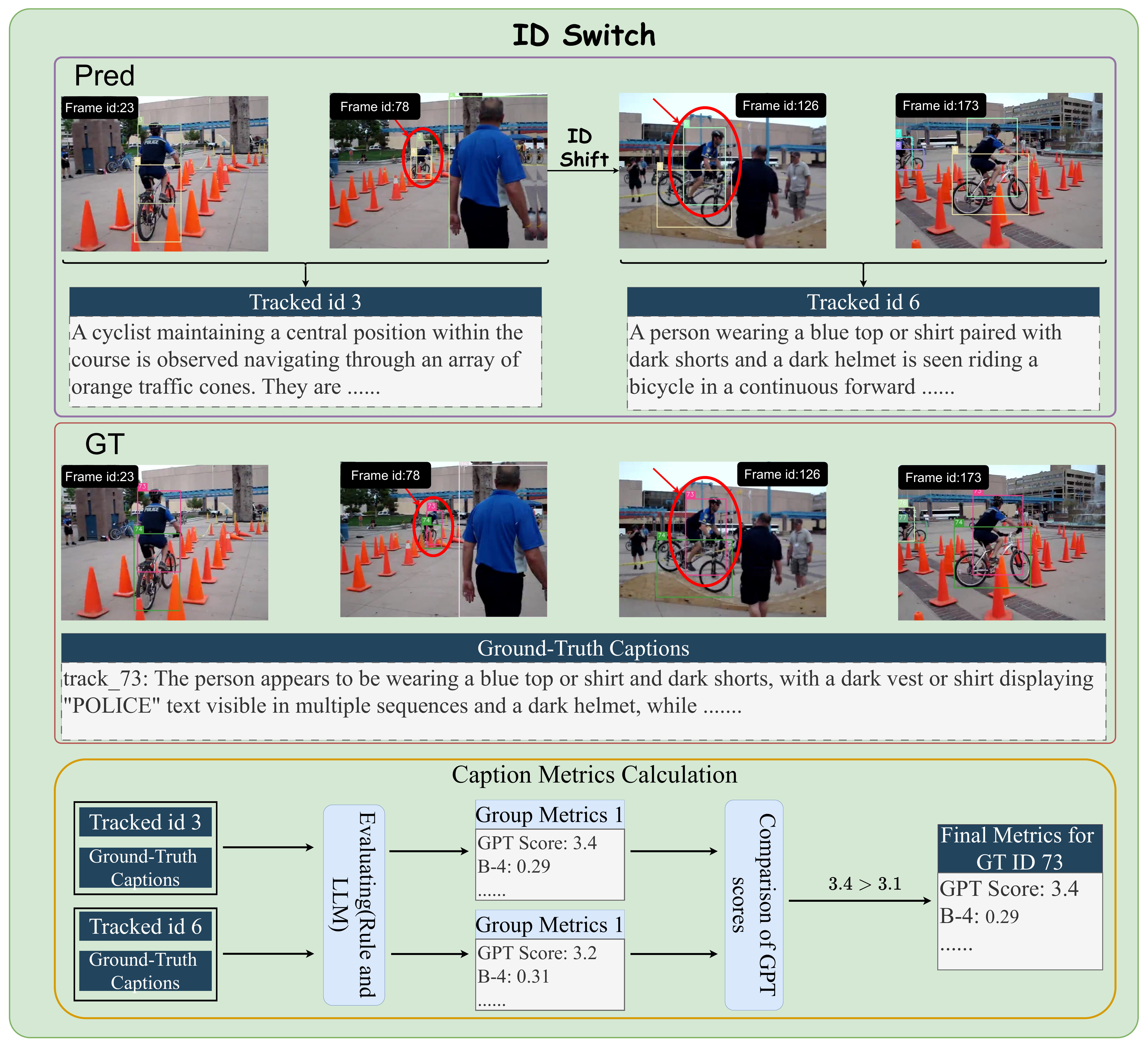}
		\caption{\textbf{Evaluation Strategy for ID Switches.} When a single ground-truth trajectory is fragmented into multiple predicted tracklets (\textit{e.g.}, Tracked ID 3 and ID 6), both predicted captions are evaluated against the GT caption. The maximum score is assigned as the final metric for the GT instance, decoupling the semantic reasoning assessment from temporal association errors.}
		\label{fig:id_switch}
		\vspace{-0.2cm}
	\end{figure}

	\textit{1) Mitigating the Impact of ID Switches:} As illustrated in Figure~\ref{fig:id_switch}, an ID switch occurs when a single Ground Truth (GT) target is fragmented into multiple predicted tracklets due to occlusion or tracking drift. To fairly assess the semantic quality of these fragmented predictions, we evaluate the generated narratives for all associated predicted tracklets $\{C_1, C_2, \dots, C_k\}$ against the unified GT caption $C_{GT}$. The final semantic score $M$ for this GT instance is defined as the maximum score achieved among its matched tracklets:
	\begin{equation}
		M(C_{GT})=\max_{i \in \{1, \dots, k\}}\text{Metric}(C_i, C_{GT})
	\end{equation}
	This operation ensures that as long as the model successfully comprehends and describes the target during at least one continuous segment of its trajectory, it receives appropriate semantic credit, while the fragmentation penalty is strictly confined to the tracking association metrics.
	
\begin{figure}[htbp]
		\centering
		\vspace{-0.1cm}
		\includegraphics[width=1\linewidth]{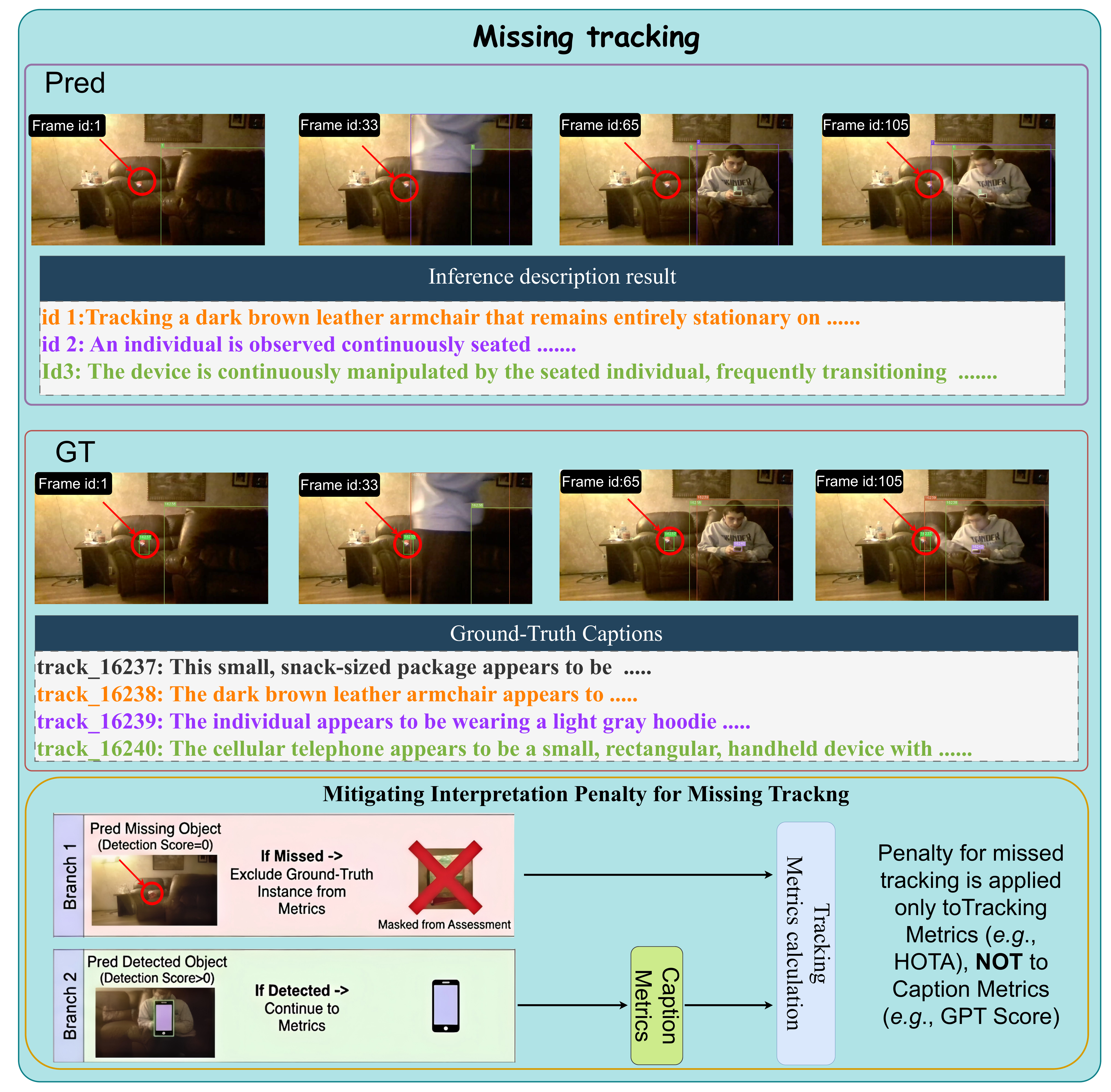}
		\caption{\textbf{Evaluation Strategy for Missing Tracking.} If a target is completely undetected by the model, it is dynamically excluded from the semantic metric calculation. This ensures that pure detection failures impact tracking metrics but do not unfairly penalize the generative understanding metrics.}
		\label{fig:missing_track}
		\vspace{-0.1cm}
	\end{figure}

	\textit{2) Masking Missing Tracks:} As shown in Figure~\ref{fig:missing_track}, severe visual occlusion or extreme scale variation can cause an object to be entirely missed by the detector (Detection Score $= 0$). To isolate the language model's descriptive capability from the detector's recall limit, we apply an Object Instance Filtering mechanism. Let $\mathcal{D}$ denote the set of successfully detected GT objects. The final corpus-level caption score is strictly computed over the detected subset $\mathcal{D}$:
	\begin{equation}
		\text{Final\_Caption\_Score}=\frac{1}{|\mathcal{D}|}\sum_{i \in \mathcal{D}}\text{Metric}(C_i, C^{GT}_i)
	\end{equation}
	By explicitly masking the missed instances from the captioning assessment, the interpretation penalty is set to zero. The failure to detect the object is rightly penalized in the geometric tracking metrics, ensuring that the semantic metrics remain a pure reflection of the model's ability to describe what it can actually ``\textit{see}".
	
	\subsection{Model Selection and Quality Assurance}

	\textbf{MLLM Selection for Generation.} Selecting an appropriate MLLM is critical for our generation pipeline. While closed-source commercial APIs (\textit{e.g.}, GPT-4o) possess formidable capabilities, they are subjected to heavy alignment taxes, resulting in formulaic prose. Moreover, their cost limits the scalability of processing long-sequence datasets like TAO. Therefore, we evaluate a wide spectrum of open-source MLLMs using an impartial GPT-4o Vision-Language Critic. To align with our dual-stream narrative structure, we evaluate models on two distinct dimensions: \textbf{GPT Video Quality Score} and \textbf{GPT Instance Quality Score} (scale 1--10). As detailed in Table~\ref{tab:mllm_ablation}, Qwen3-VL-32B achieves the highest overall quality, exceptionally balancing macro-level scene understanding with micro-level dynamic tracking, serving as our optimal generator.
	
	\begin{table}[htbp]
		\centering
		\caption{\textbf{Evaluation of Open-Source MLLMs for Narrative Generation.} Evaluated by a GPT-4o Critic (scale 1--10) on a sampled subset. The metrics are divided into Video-Level and Instance-Level Quality Scores to reflect macro and micro understanding. Qwen3-VL-32B achieves the best overall direct-generation performance.}
		\label{tab:mllm_ablation}
		\setlength{\tabcolsep}{5pt}
		\resizebox{1.0\linewidth}{!}{%
			\begin{tabular}{l|c|cc|c}
				\toprule
				\textbf{Model} & \textbf{Size} & \textbf{Video Score} $\uparrow$ & \textbf{Inst. Score} $\uparrow$ & \textbf{Average} $\uparrow$ \\
				\midrule
				MiniCPM-V 4.0~\cite{team2025minicpm4} & 4B & 6.82 & 7.65 & 7.23 \\
				LLaVA-OneVision-1.5~\cite{an2025llava} & 8B & 7.15 & 7.82 & 7.48 \\
				InternVL3.5~\cite{wang2025internvl3} & 38B & 7.55 & 8.15 & 7.85 \\
				LLaVA-OneVision~\cite{li2024llava} & 72B & 7.85 & 8.35 & 8.10 \\
				Qwen2.5-VL~\cite{wang2024qwen2}  & 72B & 8.05 & 8.52 & 8.28 \\
				\midrule
				\rowcolor{green!8}
				\textbf{Qwen3-VL}~\cite{yang2025qwen3} & 32B & 8.18 & 8.65 & 8.41 \\
				\bottomrule
			\end{tabular}%
		}
	\end{table}
	
	\textbf{Human-in-the-Loop Verification.} Despite Qwen3-VL-32B's remarkable capabilities, zero-shot generation inevitably introduces occasional hallucinations. To guarantee rigorous semantic fidelity at scale without unmanageable labor costs, we implement the automated filtering and manual verification pipeline detailed in Section~\ref{subsec:qa}. This hybrid protocol ensures human-level physical correctness while drastically optimizing overall annotation costs.

	\section{LLMtrack}
	
	\begin{figure*}[htbp]
		\centering 
		\includegraphics[width=1\textwidth]{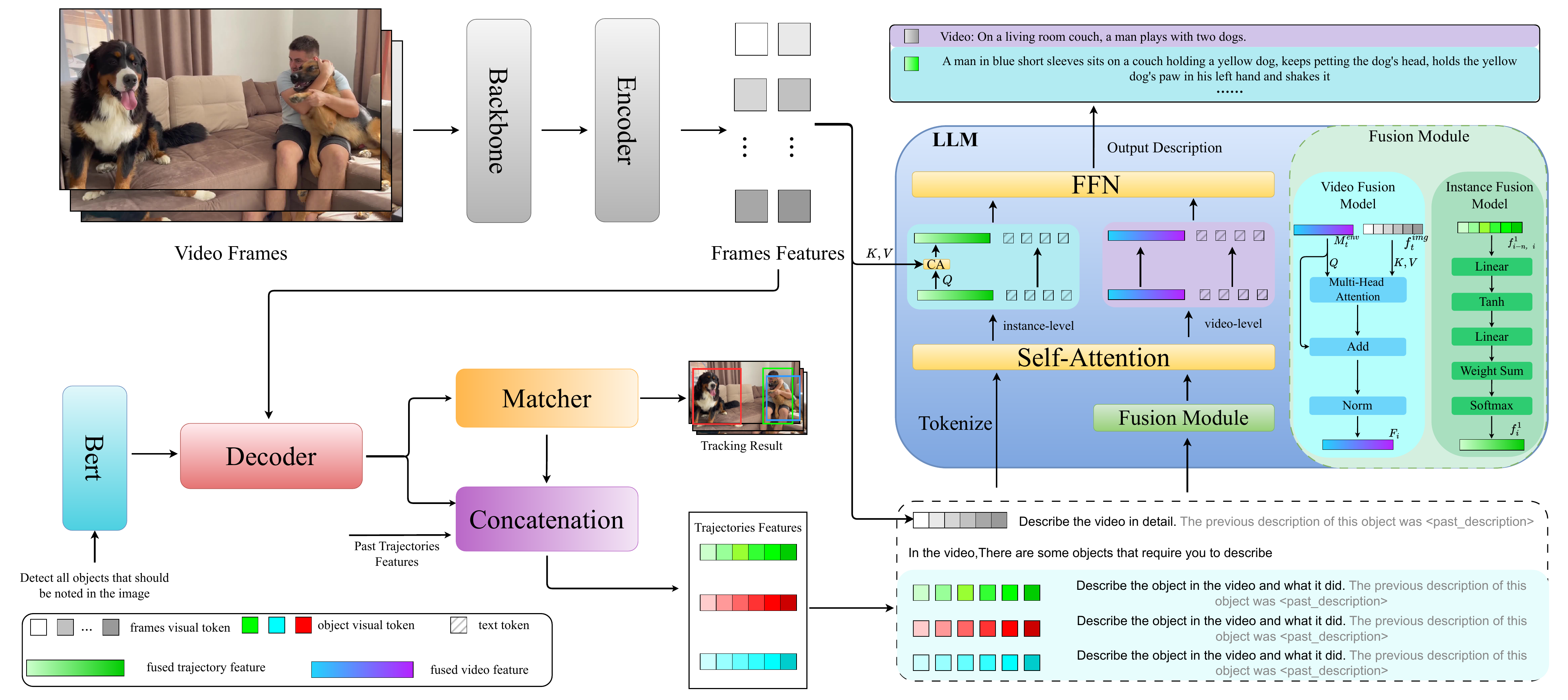} \caption{LLMTrack architecture. Online processing uses backbone visual features and tracker instance features in a Fusion Module (Video Fusion for long-term context, Instance Fusion for short-term dynamics). Fused tokens + previous-frame description enable LLM to generate current-frame narratives via Cross-Attention (CA), which integrates features during single-object captioning.} 
		\label{fig:llmtrack_arch} 
		\vspace{-0.2cm}
	\end{figure*}
	
	\subsection{Overview}
We formulate SMOT as a dual-branch online process involving continuous geometric association and semantic reasoning. Let $\mathcal{V} = \{I_1, I_2, \dots, I_T\}$ denote a streaming video sequence, where each frame $I_t \in \mathbb{R}^{H \times W \times 3}$. At each time step $t$, the system performs two concurrent tasks: (1) \textbf{Geometric Tracking}, yielding the spatial coordinates $B_t = \{b_t^i\}_{i=1}^{N_t}$ (with $b_t^i \in \mathbb{R}^4$) and identities for all active targets; and (2) \textbf{Semantic Interpretation}, generating a comprehensive state description $\mathcal{S}_t$.
	
	To achieve efficient online tracking, we adopt the Joint Detection and Embedding (JDE)~\cite{wang2020towards} paradigm for our visual frontend. As illustrated in Figure~\ref{fig:llmtrack_arch}, \textbf{LLMTrack} consists of this visual frontend, a Spatio-Temporal Fusion Module, and an LMM backend. Our spatial processing pipeline builds upon the detection architecture of LLMDet~\cite{fu2025llmdet}. To transcend this static perception for dynamic SMOT, we introduce continuous geometric tracking and a novel Spatio-Temporal Fusion Module that transforms independent frame-level detections into coherent video-level narratives. Specifically, we employ Grounding DINO~\cite{liu2024grounding} for detection; similar to transformer trackers that operate without additional queries or networks~\cite{liao2025fasttracktr}, we extract detection boxes and visual embeddings in a single forward pass, and utilize Byte~\cite{zhangByteTrackMultiobjectTracking2022} for robust data association. The resulting visual features are then compressed by our dual-stream Fusion Module into compact tokens. Finally, to maintain temporal consistency while mitigating the memory bottleneck of LLMs, we introduce a recursive prompting mechanism where the semantic state of the previous frame $\mathcal{S}_{t-1}$ serves as a linguistic prior for the current generation.
	
	\subsection{Spatio-Temporal Fusion Module}
	Directly feeding frame-wise visual tokens into an LLM is computationally prohibitive and introduces temporal redundancy. To bridge the gap between high-frequency visual signals and high-level semantic reasoning, we design a lightweight \textbf{Fusion Module}. This module compresses the visual stream into two distinct sets of compact tokens: \textit{Video Context Tokens} and \textit{Instance Dynamic Tokens}.
	
\textbf{Video Fusion: Global Context Aggregation.} To capture the evolving environmental atmosphere (\textit{e.g.}, weather changes, scene transitions), we employ a recursive update mechanism. Let $f_t^{img} \in \mathbb{R}^{H \times W \times C}$ be the visual features of the current frame. We maintain a running memory token $M_{t-1}^{env} \in \mathbb{R}^D$ representing the global context up to time $t-1$. The current environment token $M_t^{env} \in \mathbb{R}^D$ is updated via a Cross-Attention mechanism where the memory serves as the Query, and the current frame features serve as Key and Value:
	\begin{equation}
		\begin{aligned}
			M_t^{env} = \text{Norm}\big( M_{t-1}^{env} + \text{Attention}( &Q=M_{t-1}^{env}, \\
			&K=f_t^{img}, \\
			&V=f_t^{img}) \big)
		\end{aligned}
	\end{equation}
	This recurrent formulation allows the model to compress arbitrarily long video sequences into a fixed-size representation, strictly adhering to the online processing constraint.
	
\textbf{Instance Fusion: Adaptive Temporal Association.} Similarly, for individual objects, modeling actions requires analyzing motion patterns within a relevant timeframe. We apply the same sliding window strategy to object trajectories. For the $i$-th target at frame $t$, let $\mathbf{V}_t^i = \{v_{k}^i\}_{k=\max(0, t-L+1)}^{t}$ (where $v_k^i \in \mathbb{R}^D$) be the sequence of object visual embeddings within the window.

To fuse this sequence into a compact instance token $h_t^i$, we employ an adaptive additive attention mechanism. We calculate a significance weight $\alpha_k$ for each historical frame $k$ relative to the current state $v_t^i$:
\begin{equation}
	e_k = \mathbf{w}_2^\top \tanh(\mathbf{W}_1 [v_t^i; v_k^i] + b_1), \quad \alpha_k = \text{Softmax}(e_k)
\end{equation} where $\mathbf{W}_1 \in \mathbb{R}^{D \times 2D}$ and $\mathbf{w}_2 \in \mathbb{R}^D$ are learnable projection weights, and $b_1 \in \mathbb{R}^D$ is the bias. The final instance representation $h_t^i \in \mathbb{R}^D$ is obtained by the weighted aggregation of the historical features:

\begin{equation}
	h_t^i = \text{Linear}\left(\sum_{k} \alpha_k v_k^i\right)
\end{equation}

	By sharing the same window size $L$ across both video and instance modules, LLMTrack maintains a synchronized temporal receptive field, ensuring that the semantic reasoning for both the environment and agents is temporally aligned.
	
\textbf{Online Recursive Generation.} A core challenge in online semantic tracking is maintaining narrative consistency. We address this by conditioning the LLM generation on the immediate past and enforcing our \textit{Macro-Understanding-First} paradigm: a strict conditional generation hierarchy on the visual observation space within the MLLM's prompt. While the historical textual state ($\mathcal{S}_{t-1}$) serves as a temporal prior, within the current visual observation tokens the global video context token ($M_t^{env}$) must structurally precede any instance dynamic tokens ($\{h_t^i\}$). Anchoring the macro-level environment first lets the LLM's causal self-attention use the overarching scene atmosphere as a structural prior that guides subsequent micro-level instance descriptions and interaction deductions, suppressing hallucinated interactions. We thus construct the prompt $\mathcal{P}_t$ as follows:
	
	\begin{equation}
		\begin{split}
			\mathcal{P}_t = [\mathcal{I}; &\text{``Previous State:''} \oplus \mathcal{S}_{t-1}; \\
			&\text{``Current Observation:''} \oplus M_t^{env} \oplus \{h_t^i\}_{i=1}^{N_t}]
		\end{split}
	\end{equation}
	
	By prepending the global environment token $M_t^{env}$ to the instance tokens $\{h_t^i\}$, the LLM's causal self-attention forces the micro-level instance descriptions to condition on the macro-level context. The model then estimates the probability of the current description $\mathcal{S}_t$:
	\begin{equation}
		P(\mathcal{S}_t | \mathcal{V}_t) = \prod_{j=1}^{|S_t|} P(w_j | w_{<j}, \mathcal{P}_t)
	\end{equation}
	By explicitly including $\mathcal{S}_{t-1}$, the model is encouraged to generate differential descriptions, focusing on \textit{what has changed} relative to the last moment, which stabilizes the output in long-term tracking scenarios.
	
	\subsection{Progressive Three-Stage Training}
	
	\begin{figure} 
		\centering
		\includegraphics[width=\linewidth]{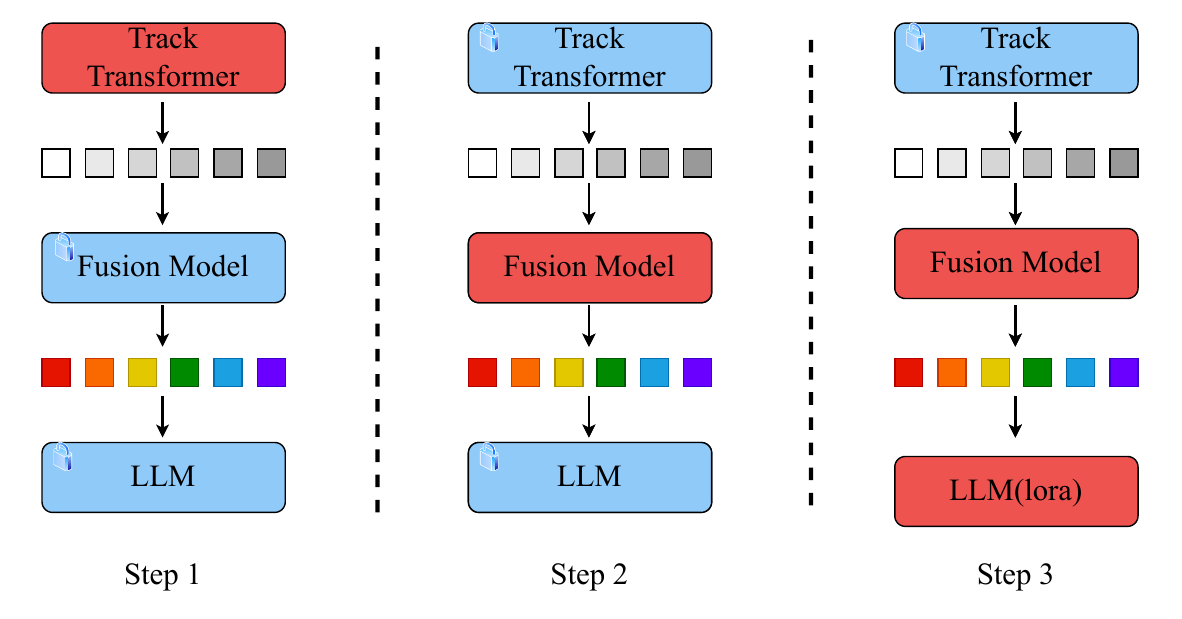}
		\caption{\textbf{The progressive three-stage training paradigm.} Red indicates trainable components, while blue indicates frozen ones. \textbf{Step 1} trains only the tracker. \textbf{Step 2} aligns the Fusion Module with the frozen LLM. \textbf{Step 3} jointly fine-tunes the Fusion Module and the LLM via LoRA~\cite{hu2022lora}.}
		\label{fig:training_step}
		\vspace{-0.2cm}
	\end{figure}
	
	Training a unified architecture for both geometric tracking and semantic reasoning is challenging. To ensure stability and efficiency, we propose a progressive training strategy (as illustrated in Figure~\ref{fig:training_step}).
	
	\noindent \textbf{Stage 1: Geometric Warm-up.}
	In the initial stage, we focus exclusively on establishing a robust geometric foundation by training only the visual tracker while keeping the Fusion Module and LLM frozen. Following the design of our prior work, FastTrackTr~\cite{liao2025fasttracktr}, we employ standard tracking losses $\mathcal{L}_{track}$. For the initial geometric tracking stage, full temporal back-propagation is impractical due to the massive scale of the visual backbone. Following the protocol in MOTIP~\cite{gao2025multiple}, we utilize a sparse frame sampling strategy (see Algorithm~\ref{alg:stage1_sparse}). 
	
	\begin{algorithm}[htbp]
		\caption{Sparse Frame Sampling Strategy for Geometric Warm-up}
		\label{alg:stage1_sparse}
		\begin{algorithmic}[1]
			\Require Video length $T$, Keyframe sample size $N$
			\State Initialize loss $\mathcal{L}_{Stage1} \leftarrow 0$
			\State Sample $N$ distinct keyframes $\mathcal{K} \subset \{1, 2, \dots, T\}$
			\For{$t = 1$ \textbf{to} $T$}
			\If{$t \in \mathcal{K}$}
			\State Enable gradient computation
			\State Extract features \& compute loss $\mathcal{L}_{track}^{(t)}$
			\State $\mathcal{L}_{Stage1} \leftarrow \mathcal{L}_{Stage1} + \mathcal{L}_{track}^{(t)}$
			\Else
			\State \texttt{\# Inference mode for motion estimation}
			\State Disable gradient computation (\texttt{no\_grad})
			\State Extract features for temporal continuity
			\EndIf
			\EndFor
			\State \textbf{Return} $\mathcal{L}_{Stage1} / N$
		\end{algorithmic}
	\end{algorithm}
	
	Crucially, this stage is strictly isolated from semantic reasoning to prevent early feature distortion. Gradients are computed only on a randomly sampled subset of key frames $\mathcal{K} = \{t_1, t_2, \dots, t_N\}$ ($N \ll T$). Intermediate frames are processed in inference mode (\texttt{no\_grad}) to maintain the temporal continuity required for motion estimation. The objective is purely geometric:
	\begin{equation}
		\mathcal{L}_{Stage1} = \frac{1}{|\mathcal{K}|} \sum_{t \in \mathcal{K}} \mathcal{L}_{track}^{(t)}
	\end{equation}
	This decouples memory usage from the video length, reducing complexity from $\mathcal{O}(T)$ to $\mathcal{O}(N)$ while establishing robust trajectory embeddings. In this paper, $N$ is 4 and $T$ is 30.
	
	\noindent \textbf{Stage 2: Semantic Alignment via Decoupled TBPTT.}
	Once the tracker converges, we freeze the tracking networks to prevent feature drift, focusing exclusively on optimizing the Fusion Module via a Causal Language Modeling (CLM) objective. Processing long sequences natively exceeds GPU memory and starves instance-level modules of gradients due to sparse, sequence-level annotations. To resolve this, we adopt a \textbf{Decoupled Dense-Sparse Optimization} strategy based on Truncated Back-Propagation Through Time (TBPTT)~\cite{weng2024longvlm,wu2022memvit}.
	
	The video $V$ is partitioned into non-overlapping training clips $\{C_1, C_2, \dots, C_M\}$. To strictly preserve the global context and adhere to our recursive update mechanism, a memory cache $\mathcal{M}$ (containing the environment token $\mathbf{H}^{global}$ and instance states) is passed sequentially from $C_{k-1}$ to $C_k$, applying a \textbf{stop-gradient operation} ($\text{sg}(\cdot)$) to bound the backward pass while preserving global history in the forward pass:
	\begin{equation}
		\mathbf{H}_k^{global}, \{\tilde{\mathbf{f}}_k^i\} = \text{Fusion}(C_k \mid \text{sg}(\mathbf{H}_{k-1}^{global}), \text{sg}(\{\tilde{\mathbf{f}}_{k-1}^i\}))
	\end{equation}
	
	To handle sparse annotations in long sequences, we employ an extrapolation strategy that decouples spatial association from the disappearance mechanism. Let $t$ denote the current frame and $\tau$ the nearest annotated frame. Given predicted bounding boxes $b_t^i$ and semantic grounding scores $s_t^i$, the target assignment $y_t^i$ is defined as:
	\begin{equation}
		y_t^i = \begin{cases}
			\arg\max_{j} \text{IoU}(b_t^i, \hat{b}_\tau^j), & \text{if } s_t^i \geq \theta_{score} \\
			& \text{and } \max_{j} \text{IoU}(b_t^i, \hat{b}_\tau^j) \geq \theta_{IoU} \\[1ex]
			\varnothing, & \text{otherwise}
		\end{cases}
	\end{equation}
	If the grounding confidence falls below $\theta_{score}$, it is immediately assigned to the null state ($\varnothing$), effectively modeling object disappearance. We decouple the CLM objective to match the temporal granularity of the annotations:
	\begin{equation}
		\label{equ:loss}
		\mathcal{L}_{Stage2}^{(k)} =
		\begin{cases}
			\mathcal{L}_{instance}(C_k) & \text{if } k < K \\
			\mathcal{L}_{instance}(C_K) + \lambda_{vid} \mathcal{L}_{video}(\mathbf{H}_K^{global}) & \text{if } k = K
		\end{cases}
	\end{equation}
	Here, $\mathcal{L}_{instance}$ acts as a \textbf{dense supervisor} computed at every clip to continuously optimize fine-grained trajectory features, while $\mathcal{L}_{video}$ serves as a \textbf{sparse supervisor}, activated exclusively at the final clip $K$ where the environment token $\mathbf{H}_K^{global}$ has fully encapsulated the holistic scene narrative.
	
	\noindent \textbf{Stage 3: Cognitive Fine-tuning via LoRA.}
	In the final stage, we keep the visual tracker frozen while jointly fine-tuning the aligned Fusion Module and the LLM through Low-Rank Adaptation (LoRA)~\cite{hu2022lora}, continuing to adhere to the Decoupled TBPTT protocol established in Stage 2. Joint optimization allows the Fusion Module to adapt its spatio-temporal representations to the activated LLM while preserving the geometric features learned by the frozen tracker. Although gradients are temporally truncated between consecutive clips, the model learns to generate temporally coherent narratives by conditioning on both the accumulated textual history embedded in the prompts and the progressively adapted visual tokens. The optimization objective remains the decoupled Causal Language Modeling loss defined in Eq.~\ref{equ:loss}.
	
\subsection{Paradigm Shift in Interaction Modeling}
\label{sec:paradigm_shift}
As posited in the Introduction, a core philosophy of our work is that multi-object ``\textit{interaction}'' should not be treated as an isolated visual recognition task, but rather as a natural logical deduction that emerges from the interplay between individual dynamic behaviors and the macroscopic environmental context.

\begin{figure}[htbp]
	\vspace{-0.2cm}
	\centering
	\includegraphics[width=1\linewidth]{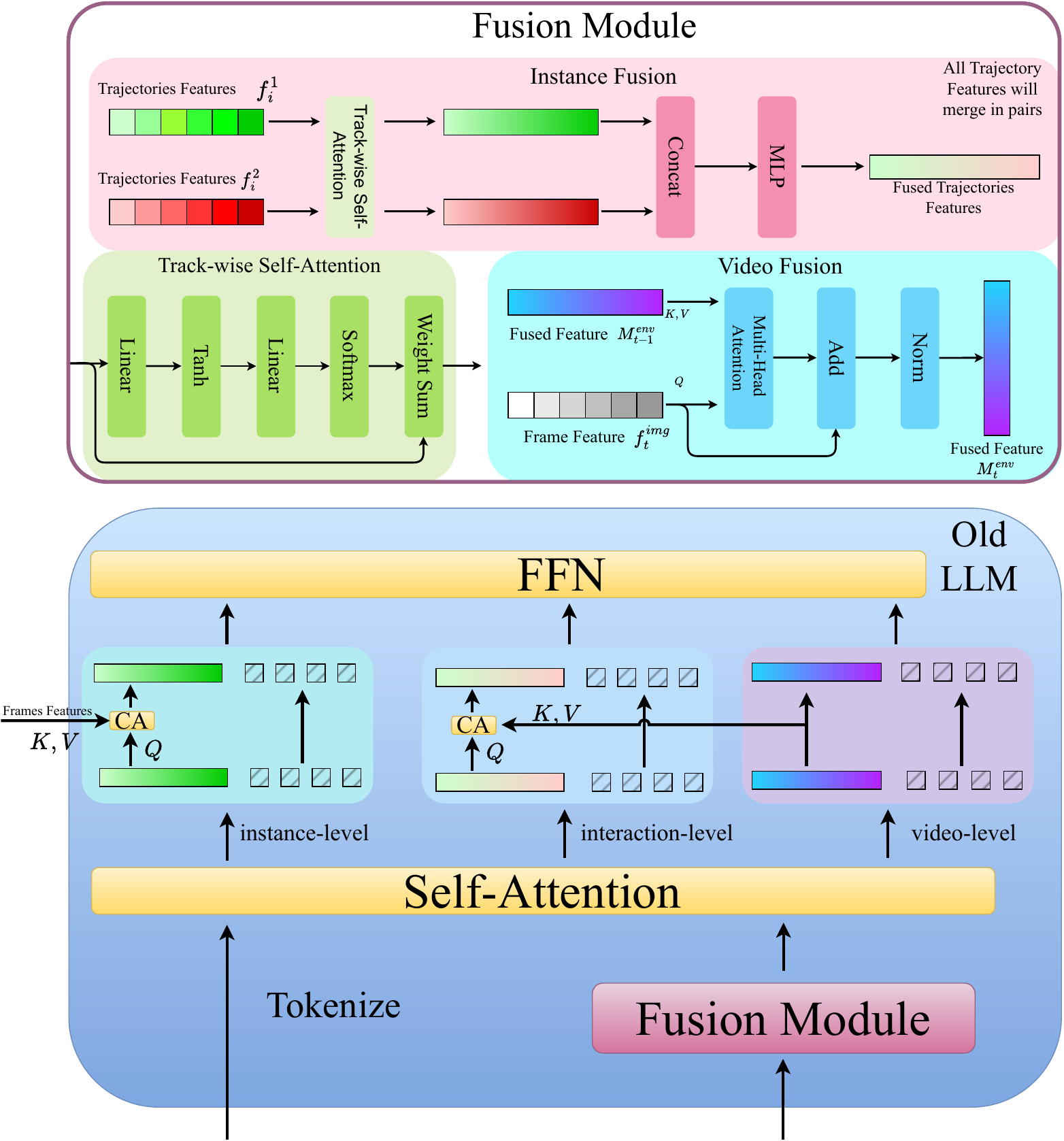}
	\caption{The detailed architecture of our early explicit feature fusion design. This complex architectural exploration serves as the explicit visual modeling baseline compared against our zero-shot text deduction paradigm.}
	\label{fig:fusion_module_appendix}
	\vspace{-0.2cm}
\end{figure}

To empirically validate this hypothesis, we initially designed an explicit interaction-level feature fusion branch as a baseline. Figure~\ref{fig:fusion_module_appendix} illustrates this architecture, which explicitly models instance, interaction, and video contexts. The trajectory features are merged in pairs via a Multi-Layer Perceptron (MLP) to yield the fused interaction trajectory feature: $F_{\text{inter}} = \text{MLP}([\tilde{f}_i^1 ; \tilde{f}_i^2])$, and fed into the LLM, where a dedicated Cross-Attention mechanism aligns the interaction tokens with the video tokens prior to the Feed-Forward Networks.

While traditional paradigms design such explicit feature fusion modules, we hypothesize that this elaborate architectural engineering is semantically redundant: if a model possesses sufficient macro-understanding and micro-tracking capabilities, complex social interactions should naturally emerge from the textual narrative. To verify this, we designed a zero-shot text-only deduction experiment that entirely discards the explicit visual interaction features and relies exclusively on the logical reasoning of the LLM. In this setting, the model deduces interactions implicitly from the foundational environment and individual dynamic tokens, guided solely by our Macro-Understanding-First prompt structure.
	
\section{Experiments}

\subsection{Experimental Setup}

\textbf{Datasets and Metrics.} We conduct evaluations on the Grand-SMOT benchmark, adhering to the official splits of its constituent sources (TAO~\cite{dave2020tao} and BenSMOT~\cite{li2024beyond}). For the TAO subset, which emphasizes open-vocabulary classification, we employ the \textbf{TETA} protocol~\cite{li2022tracking} to decompose performance into localization (LocA), association (AssocA), and classification (ClsA). For the BenSMOT subset, we report standard HOTA, DetA, AssA, and IDF1 metrics~\cite{luiten2021hota}. To evaluate interaction deduction, we adopt standard Precision, Recall, and F1 scores. For semantic reasoning, we utilize N-gram metrics (BLEU~\cite{papineni2002bleu}, METEOR, CIDEr~\cite{vedantam2015cider}) and introduce an \textbf{LLM-as-a-Judge Committee Semantic Score (GPT-S)} (1-5 Likert) that employs leading foundation models to capture logical coherence, temporal consistency, and narrative depth. Given the prohibitive API costs and latency of deploying an ensemble of SOTA commercial LLMs, the GPT-S is conducted on a uniformly sampled subset of the test set (\textit{e.g}., $N=200$ sequences), maintaining identical long-tail category distributions and sequence length variances to the full test set for statistical significance and fair comparison.

\textbf{Implementation Details.} We instantiate our visual frontend using the Grounding DINO-T architecture~\cite{liu2024grounding}, which employs a Swin-Transformer-Tiny \cite{liu2021swin} backbone. For the cognitive backend, we integrate this perception module with LLaVA-OneVision~\cite{li2024llava} and LLaVA-OneVision-1.5~\cite{an2025llava}. To facilitate efficient parameter tuning, we apply ~\cite{hu2022lora} with a rank of $r=64$. Training is executed on a node with $8 \times$ NVIDIA RTX 5880 Ada GPUs. In the ablation experiment, unless otherwise stated, training and testing were conducted only on the BenSMOT part. The first stage was trained for 12 epochs, while the second and third stages were both trained for 6 epochs. During inference on a single RTX 5880 Ada GPU, the LLMTrack-0.5B variant achieves near real-time processing at approximately 21 FPS, while the more capable LLMTrack-4B operates at roughly 13 FPS, balancing macro-cognitive reasoning depth with practical inference latency.

\subsection{Main Results}

\textbf{System-Level Tracking.} Table~\ref{tab:main_tracking} evaluates system-level tracking performance against representative recent approaches, including the 2025 ID-prediction tracker MOTIP~\cite{gao2025multiple}, the unknown-category tracking method AED~\cite{fang2025associate}, and OVTR~\cite{li2025ovtr}. LLMTrack achieves 75.23\% HOTA on BenSMOT, surpassing OC-SORT (71.74\%). To reflect peak practical capabilities, traditional baselines are evaluated using their officially optimized detectors, while LLMTrack intrinsically relies on Grounding DINO, highlighting the architectural advantage of our unified open-world pipeline over traditional closed-set systems. For the TAO Open-World Subset, we adopt TETA instead of HOTA to prevent rare-class classification errors from masking association quality under severe long-tail distributions. While LLMTrack excels in localization (LocA), its classification accuracy (ClsA) on TAO's extreme long-tail categories slightly trails OVTrack---an expected trade-off, as our model optimizes for generative sequence reasoning rather than being fine-tuned on TAO's isolated rare-class taxonomy.

\begin{table*}[htbp]
    \centering
    \caption{\textbf{Tracking performance on the Grand-SMOT benchmark.} This includes a \textbf{Standard Subset} (BenSMOT~\cite{li2024beyond}) and an \textbf{Open-World Subset} (TAO~\cite{dave2020tao}). Best in \best{red}, second in \second{blue}. Methods with an asterisk (*) use their officially optimized detectors (Faster R-CNN~\cite{ren2016faster} for TAO; YOLOX~\cite{ge2021yolox} for BenSMOT) to ensure fair evaluation at peak capacity, whereas LLMTrack leverages Grounding DINO to demonstrate a unified open-world paradigm. ``--'' denotes a method not yet reproduced under the Grand-SMOT protocol. To rigorously decouple association capability from detector performance, Table~\ref{tab:controlled_association} provides a controlled experiment using identical Grounding DINO detections.}
    \label{tab:main_tracking}
    \resizebox{!}{!}{%
        \begin{tabular}{l|cccc|cccc}
            \toprule
            \multirow{2}{*}{\textbf{Method}} & \multicolumn{4}{c|}{\textbf{Grand-SMOT (BenSMOT Split)}} & \multicolumn{4}{c}{\textbf{Grand-SMOT (TAO Split)}} \\
            \cmidrule(lr){2-5} \cmidrule(lr){6-9}
            & \textbf{HOTA} $\uparrow$ & \textbf{DetA} $\uparrow$ & \textbf{AssA} $\uparrow$ & \textbf{IDF1} $\uparrow$ & \textbf{TETA} $\uparrow$ & \textbf{LocA} $\uparrow$ & \textbf{AssocA} $\uparrow$ & \textbf{ClsA} $\uparrow$ \\
            \midrule
            SORT* \cite{bewleySimpleOnlineRealtime2016} & 48.8 & 60.5 & 39.2 & 53.5 & 24.8 & 48.1 & 14.3 & 12.1 \\
            DeepSORT* \cite{wojke2017simple} & 49.8 & 61.9 & 39.8 & 57.2 & 26.0 & 48.4 & 17.5 & 12.1 \\
            ByteTrack* \cite{zhangByteTrackMultiobjectTracking2022} & 68.2 & 67.6 & 70.8 & 78.9 & 29.2 & 49.1 & 26.8 & 12.1 \\
            OC-SORT* \cite{cao2023observation} & 71.7 & 69.1 & 73.6 & 78.2 & 27.5 & 48.5 & 22.1 & 12.1 \\
            Hybrid-SORT* \cite{yang2024hybrid} & 72.2 & 69.3 & 75.2 & 79.4 & 28.1 & 48.8 & 23.5 & 12.1 \\
            OVTR \cite{li2025ovtr} & 68.4 & 63.5 & 73.7 & 71.5 & 33.2 & \second{50.0} & 34.5 & 15.1 \\
            MOTIP \cite{gao2025multiple} & 73.1 & 70.1 & \second{76.3} & 80.1 & 30.5 & 49.0 & 28.5 & 12.1 \\
            AED* \cite{fang2025associate} & 71.8 & 65.8 & 76.0 & 73.9 & 34.2 & 49.3 & 35.5 & \second{16.5} \\
            OVTrack* \cite{li2023ovtrack} & 69.2 & 64.2 & 74.5 & 72.4 & \best{34.7} & 49.3 & \best{36.7} & \best{18.1} \\
            SMOTer \cite{li2024beyond} & 71.6 & \second{71.3} & 73.3 & \second{80.2} & - & - & - & - \\
            \midrule
            \rowcolor{green!8}
            \textbf{LLMTrack (Ours)} & \best{75.2} & \best{73.7} & \best{77.5} & \best{84.3} & \second{34.3} & \best{51.1} & \second{36.5} & 15.0 \\
            \bottomrule
        \end{tabular}
    }
\end{table*}

\textbf{Semantic Understanding.} To address concerns regarding the source of performance gains, we establish a controlled evaluation protocol: we equip all traditional tracking-by-detection baselines (\textit{e.g.}, ByteTrack, DeepSORT) with our exact Spatio-Temporal Fusion Module and the identical 0.5B LLM backend. Recent trackers whose Grand-SMOT results are not yet reproduced are therefore evaluated only in the tracking protocol of Table~\ref{tab:main_tracking}; extending the semantic protocol requires their trajectory outputs under the same detector and LLM interface. Specifically, we utilize the officially optimized tracking bounding boxes to crop features from a frozen Swin-Transformer-Tiny via RoIAlign.

As shown in Table~\ref{tab:main_understanding}, even with the same LLM backend, LLMTrack-0.5B consistently prevails across all metrics, proving that our natively extracted, continuous trajectory representations capture far richer semantic context than standard disjointed bounding-box crops. Furthermore, LLMTrack-4B achieves a qualitative leap, reaching a Video CIDEr of 0.425 and an average GPT-S of 3.8 on BenSMOT.

\begin{table*}[htbp]
    \centering
    \caption{\textbf{Semantic Understanding performance on Grand-SMOT.} To ensure a strictly fair comparison, methods marked with an asterisk (*) utilize the exact same 0.5B LLM backend and fusion module as LLMTrack-0.5B. \textbf{GPT-S} denotes the overall Semantic Score (1-5), which is averaged over five fine-grained evaluation dimensions. Detailed dimension-wise results are provided in the \textbf{Appendix}.}
    \label{tab:main_understanding}
    \setlength{\tabcolsep}{2.5pt} 
    \resizebox{!}{!}{%
        \begin{tabular}{l|cccc|cccc|cccc|cccc}
            \toprule
            \multirow{3}{*}{\textbf{Method}} & \multicolumn{8}{c|}{\textbf{Grand-SMOT (BenSMOT Split)}} & \multicolumn{8}{c}{\textbf{Grand-SMOT (TAO Split)}} \\
            \cmidrule(lr){2-9} \cmidrule(lr){10-17}
            & \multicolumn{4}{c|}{\textit{Video Caption}} & \multicolumn{4}{c|}{\textit{Instance Caption}} & \multicolumn{4}{c|}{\textit{Video Caption}} & \multicolumn{4}{c}{\textit{Instance Caption}} \\
            \cmidrule(lr){2-5} \cmidrule(lr){6-9} \cmidrule(lr){10-13} \cmidrule(lr){14-17}
            & \textbf{B-4}$\uparrow$ & \textbf{M}$\uparrow$ & \textbf{C}$\uparrow$ & \textbf{GPT-S}$\uparrow$ & \textbf{B-4}$\uparrow$ & \textbf{M}$\uparrow$ & \textbf{C}$\uparrow$ & \textbf{GPT-S}$\uparrow$ & \textbf{B-4}$\uparrow$ & \textbf{M}$\uparrow$ & \textbf{C}$\uparrow$ & \textbf{GPT-S}$\uparrow$ & \textbf{B-4}$\uparrow$ & \textbf{M}$\uparrow$ & \textbf{C}$\uparrow$ & \textbf{GPT-S}$\uparrow$ \\
            \midrule
            SORT* \cite{bewleySimpleOnlineRealtime2016} & 0.115 & 0.141 & 0.225 & 2.70 & 0.132 & 0.155 & 0.250 & 1.40 & 0.102 & 0.130 & 0.195 & 2.48 & 0.118 & 0.135 & 0.215 & 1.53 \\
            DeepSORT* \cite{wojke2017simple} & 0.122 & 0.148 & 0.238 & 2.80 & 0.145 & 0.162 & 0.265 & 1.50 & 0.108 & 0.135 & 0.208 & 2.63 & 0.125 & 0.142 & 0.228 & 1.58 \\
            ByteTrack* \cite{zhangByteTrackMultiobjectTracking2022} & 0.142 & 0.165 & 0.270 & 3.00 & 0.168 & 0.185 & 0.305 & 1.70 & 0.125 & 0.150 & 0.235 & 2.80 & 0.145 & 0.162 & 0.260 & 1.78 \\
            OC-SORT* \cite{cao2023observation} & 0.148 & 0.168 & 0.285 & 3.10 & 0.172 & 0.192 & 0.320 & 1.80 & 0.128 & 0.152 & 0.245 & 2.93 & 0.148 & 0.168 & 0.268 & 1.88 \\
            Hybrid-SORT* \cite{yang2024hybrid} & 0.150 & 0.172 & 0.295 & 3.10 & 0.176 & 0.195 & 0.335 & 1.80 & 0.130 & 0.158 & 0.255 & 2.98 & 0.150 & 0.172 & 0.282 & 1.93 \\
            AED* \cite{fang2025associate} & 0.151 & 0.176 & 0.302 & 3.12 & 0.183 & 0.196 & 0.347 & 1.84 & 0.136 & 0.160 & 0.275 & 3.09 & 0.150 & 0.174 & 0.295 & 1.99 \\
            OVTrack* \cite{li2023ovtrack} & 0.152 & 0.178 & 0.305 & 3.10 & \second{0.184} & 0.198 & 0.350 & 1.88 & \second{0.138} & 0.162 & \second{0.278} & 3.12 & 0.152 & 0.176 & 0.298 & 1.98 \\
            MOTIP \cite{gao2025multiple} & 0.155 & 0.180 & 0.310 & 3.18 & 0.181 & 0.202 & 0.352 & 1.90 & 0.134 & 0.163 & 0.273 & 3.15 & 0.154 & 0.178 & 0.300 & 2.04 \\
            OVTR \cite{li2025ovtr} & \second{0.160} & 0.182 & \second{0.318} & 3.20 & 0.180 & \second{0.208} & 0.348 & 1.90 & 0.132 & \second{0.166} & 0.270 & 3.13 & \second{0.158} & 0.180 & 0.302 & 2.03 \\
			SMOTer \cite{li2024beyond} & 0.045 & 0.095 & 0.110 & 2.10 & 0.062 & 0.112 & 0.145 & 1.10 & - & - & - & - & - & - & - & - \\
            
			\midrule
            \rowcolor{green!8}
            \textbf{LLMTrack-0.5B (Ours)} & 0.156 & \second{0.185} & 0.315 & \second{3.30} & 0.182 & 0.205 & \second{0.355} & \second{2.00} & 0.135 & 0.165 & 0.275 & \second{3.23} & 0.155 & \second{0.182} & \second{0.305} & \second{2.13} \\
            \rowcolor{green!12}
            \textbf{LLMTrack-4B (Ours)} & \best{0.198} & \best{0.224} & \best{0.425} & \best{3.80} & \best{0.235} & \best{0.248} & \best{0.485} & \best{2.90} & \best{0.175} & \best{0.198} & \best{0.380} & \best{3.88} & \best{0.195} & \best{0.215} & \best{0.415} & \best{2.85} \\
            \bottomrule
        \end{tabular}%
    }
    \vspace{-0.2cm}
\end{table*}

\section{Ablation and Analysis}

\subsection{Ablation Studies}
\noindent\textbf{Experimental Setup.} To conserve computational resources, all ablation studies are conducted exclusively on the \textbf{BenSMOT} dataset utilizing the \textbf{LLMTrack-0.5B} model variant.

\textbf{Component Analysis.} We validate the contribution of each core component in Table~\ref{tab:ablation_component}. The addition of Grounding DINO (GD) immediately improves HOTA. However, semantic metrics only show significant improvement after introducing LoRA. Most importantly, the synergistic combination of Instance Fusion (Ins-Fus) and Video Fusion (Vid-Fus) proves essential for capturing interactions and summarizing the scene, yielding the highest Video CIDEr score.

\begin{table}[htbp]
    \centering
    \caption{\textbf{Ablation study on core components and data scale.} We progressively integrate modules and data sources to validate their contributions. \textbf{Data}: Training source (Ben: BenSMOT only, +TAO: BenSMOT + TAO); \textbf{GD}: Grounding DINO frontend; \textbf{LoRA}: Low-Rank Adaptation for LLM; \textbf{Vid}: Video Fusion Module; \textbf{Ins}: Instance Fusion Module.}
    \label{tab:ablation_component}
    \setlength{\tabcolsep}{2.5pt} 
    \resizebox{1.0\linewidth}{!}{%
        \begin{tabular}{c|cccc|cc|ccc|ccc}
            \toprule
            \multirow{2}{*}{\textbf{Data}} & \multicolumn{4}{c|}{\textbf{Components}} & \multicolumn{2}{c|}{\textbf{Tracking}} & \multicolumn{3}{c|}{\textbf{Video Summary}} & \multicolumn{3}{c}{\textbf{Instance Desc.}} \\
            \cmidrule(lr){2-5} \cmidrule(lr){6-7} \cmidrule(lr){8-10} \cmidrule(lr){11-13}
            & \textbf{GD} & \textbf{LoRA} & \textbf{Vid} & \textbf{Ins} & \textbf{HOTA} & \textbf{IDF1} & \textbf{METEOR} & \textbf{CIDEr} & \textbf{GPT-S} & \textbf{METEOR} & \textbf{CIDEr} & \textbf{GPT-S} \\
            \midrule
            Ben & - & - & - & - & 72.58 & 81.68 & 0.084 & 0.108 & 1.88 & 0.095 & 0.121 & 1.08 \\
            Ben & \checkmark & - & - & - & 74.61 & 83.52 & \textcolor{gray}{0.086} & \textcolor{gray}{0.112} & \textcolor{gray}{1.92} & \textcolor{gray}{0.098} & \textcolor{gray}{0.125} & \textcolor{gray}{1.12} \\
            Ben & \checkmark & \checkmark & - & - & \textcolor{gray}{74.61} & \textcolor{gray}{83.52} & 0.148 & 0.236 & 2.58 & 0.162 & 0.248 & 1.48 \\
            Ben & \checkmark & \checkmark & \checkmark & - & \textcolor{gray}{74.61} & \textcolor{gray}{83.52} & 0.175 & 0.292 & 2.98 & \textcolor{gray}{0.165} & \textcolor{gray}{0.252} & \textcolor{gray}{1.53} \\
            Ben & \checkmark & \checkmark & \checkmark & \checkmark & \textcolor{gray}{74.61} & \textcolor{gray}{83.52} & \textcolor{gray}{0.178} & \textcolor{gray}{0.298} & \textcolor{gray}{3.12} & 0.195 & 0.338 & 1.88 \\
            \midrule
            \rowcolor{green!8} 
            +TAO & \checkmark & \checkmark & \checkmark & \checkmark & \textbf{75.23} & \textbf{84.31} & \textbf{0.185} & \textbf{0.315} & \textbf{3.35} & \textbf{0.205} & \textbf{0.355} & \textbf{2.05} \\
            \bottomrule
        \end{tabular}%
    }
\end{table}

\textbf{Progressive Training Paradigms.} We validate our three-stage pipeline in Table~\ref{tab:ablation_training_llm}. Skipping the Stage 2 alignment drastically degrades generation quality (CIDEr drops to 0.245), proving that a shared latent space is essential prior to LLM activation. Furthermore, freezing the Fusion Module during Stage 3 yields suboptimal temporal representations. Jointly fine-tuning it alongside the LLM is therefore crucial for optimal performance.

\begin{table}[htbp]
    \centering
    \caption{\textbf{Impact of different training paradigms on model convergence and performance.} We analyze the contribution of the intermediate alignment stage and the necessity of tuning the fusion module during the final stage. Tracking performance remains consistent across all variants and is omitted for brevity.}
    \label{tab:ablation_training_llm}
    \vspace{-5pt}
    \resizebox{1\linewidth}{!}{%
        \begin{tabular}{l|ccc|ccc}
            \toprule
            \multirow{2}{*}{\textbf{Training Strategy}} & \multicolumn{3}{c|}{\textbf{Instance Desc.}} & \multicolumn{3}{c}{\textbf{Video Summary}} \\
            \cmidrule(lr){2-4} \cmidrule(lr){5-7}
            & \textbf{METEOR} $\uparrow$ & \textbf{CIDEr} $\uparrow$ & \textbf{GPT-S} $\uparrow$ & \textbf{METEOR} $\uparrow$ & \textbf{CIDEr} $\uparrow$ & \textbf{GPT-S} $\uparrow$ \\
            \midrule
            Stage 1 $\rightarrow$ Stage 3 (w/o Stage 2) & 0.160 & 0.245 & 1.45 & 0.145 & 0.202 & 2.38 \\
            Stage 1 $\rightarrow$ Stage 2 $\rightarrow$ Stage 3 (w/o Fusion) & 0.188 & 0.320 & 1.78 & 0.172 & 0.285 & 2.92 \\
            \midrule
            \rowcolor{green!8}
            \textbf{Stage 1 $\rightarrow$ Stage 2 $\rightarrow$ Stage 3 (Full)} & \textbf{0.195} & \textbf{0.338} & \textbf{1.88} & \textbf{0.178} & \textbf{0.298} & \textbf{3.12} \\
            \bottomrule
        \end{tabular}%
    }
\end{table}

\textbf{Temporal Fusion Window Size.} Table~\ref{tab:ablation_frames} ablates the temporal aggregation window size ($L$). A restricted window ($L=10$) fails to capture complex actions, severely reducing CIDEr. Expanding to $L=30$ optimally balances context retention and online processing constraints, achieving results highly competitive with the offline upper bound ($L=\text{Full sequence}$). Thus, we adopt $L=30$ as our default setting.

\begin{table}[htbp]
    \centering
    \caption{\textbf{Impact of fusion window size ($L$).} We evaluate the temporal aggregation range on the BenSMOT split (0.5B model). $L=\text{Full}$ serves as an offline upper bound. Notably, $L=\text{Full}$ is only feasible for BenSMOT's short sequences ($\sim$70 frames), as TAO's lengthy sequences ($>$1000 frames) make full-sequence processing computationally intractable.}
    \vspace{-5pt}
    \label{tab:ablation_frames}
    \setlength{\tabcolsep}{8pt}
    \resizebox{1\linewidth}{!}{%
        \begin{tabular}{l|ccc|ccc}
            \toprule
            \multirow{2}{*}{\textbf{Window Size ($L$)}} & \multicolumn{3}{c|}{\textbf{Instance Desc.}} & \multicolumn{3}{c}{\textbf{Video Summary}} \\
            \cmidrule(lr){2-4} \cmidrule(lr){5-7} 
            & \textbf{METEOR} $\uparrow$ & \textbf{CIDEr} $\uparrow$ & \textbf{GPT-S} $\uparrow$ & \textbf{METEOR} $\uparrow$ & \textbf{CIDEr} $\uparrow$ & \textbf{GPT-S} $\uparrow$ \\
            \midrule
            $L=10$ (Short) & 0.175 & 0.284 & 1.58 & 0.160 & 0.256 & 2.68 \\
            $L=20$ (Medium) & 0.188 & 0.327 & 1.82 & 0.171 & 0.293 & 3.03 \\
            \rowcolor{green!8}
            \textbf{$L=30$ (Default Online)} & \textbf{0.195} & \textbf{0.338} & \textbf{1.88} & \textbf{0.178} & \textbf{0.298} & \textbf{3.12} \\
            \midrule
            \textcolor{gray}{$L=$ Full (Offline Bound)} & \textcolor{gray}{0.194} & \textcolor{gray}{0.341} & \textcolor{gray}{1.92} & \textcolor{gray}{0.180} & \textcolor{gray}{0.320} & \textcolor{gray}{3.38} \\
            \bottomrule
        \end{tabular}%
    }
\end{table}

\textbf{Effectiveness of the Annotation Pipeline.} To validate our annotation strategy, we assess datasets generated under different configurations (Table~\ref{tab:ablation_generation}) using the GPT-4o Critic on a 470-sequence subset. Pure manual annotation yields lower overall scores, as human annotators struggle to consistently produce dense, micro-level narratives. Direct MLLM generation improves textual richness but introduces spatial-temporal hallucinations. While unconstrained manual correction of MLLM outputs achieves high textual quality, it often misses underlying physical grounding errors. Our standard pipeline resolves this by integrating Qwen3-VL with an automated Vision-Language Critic and targeted Human-in-the-Loop refinement, optimally balancing narrative depth with factual alignment and eliminating hallucinations.

\begin{table}[htbp]
	\centering
	\caption{\textbf{Evaluation of Generation Strategies Across MLLMs.} We assess generation quality using a GPT-4o Critic (scale 1--10) on Video and Instance Quality. Compared to manual annotation (BenSMOT baseline), direct generation yields lower fidelity, and unconstrained manual extension lacks consistency despite occasional peaks. }
	\label{tab:ablation_generation}
	\setlength{\tabcolsep}{4pt}
	\resizebox{1.0\linewidth}{!}{%
		\begin{tabular}{l|c|l|cc|c}
			\toprule
			\textbf{Base Model} & \textbf{Size} & \textbf{Generation Strategy} & \textbf{Video Score} $\uparrow$ & \textbf{Inst. Score} $\uparrow$ & \textbf{Average} $\uparrow$ \\
			\midrule
			\textbf{Human Baseline} & - & Pure Manual Annotation & 7.21 & 7.08 & 7.15 \\
			\midrule
			\multirow{3}{*}{MiniCPM-V 4.0~\cite{team2025minicpm4}} & \multirow{3}{*}{4B}
			& Direct Generation & 6.82 & 7.65 & 7.23 \\
			& & Full Manual Extension & 7.15 & 7.95 & 7.55 \\
			& & \textbf{Ours (Pipeline + QA)} & 7.25 & 8.02 & 7.63 \\
			\midrule
			\multirow{3}{*}{LLaVA-OneVision-1.5~\cite{an2025llava}} & \multirow{3}{*}{8B}
			& Direct Generation & 7.15 & 7.82 & 7.48 \\
			& & Full Manual Extension & 7.45 & 8.12 & 7.78 \\
			& & \textbf{Ours (Pipeline + QA)} & 7.58 & 8.20 & 7.89 \\
			\midrule
			\multirow{3}{*}{InternVL3.5~\cite{wang2025internvl3}} & \multirow{3}{*}{38B}
			& Direct Generation & 7.55 & 8.15 & 7.85 \\
			& & Full Manual Extension & 7.92 & 8.42 & 8.17 \\
			& & \textbf{Ours (Pipeline + QA)} & 7.85 & 8.55 & 8.20 \\
			\midrule
			\multirow{3}{*}{LLaVA-OneVision~\cite{li2024llava}} & \multirow{3}{*}{72B}
			& Direct Generation & 7.85 & 8.35 & 8.10 \\
			& & Full Manual Extension & 8.25 & 8.65 & 8.45 \\
			& & \textbf{Ours (Pipeline + QA)} & 8.15 & 8.78 & 8.46 \\
			\midrule
			\multirow{3}{*}{Qwen2.5-VL~\cite{wang2024qwen2}} & \multirow{3}{*}{72B}
			& Direct Generation & 8.05 & 8.52 & 8.28 \\
			& & Full Manual Extension & 8.42 & 8.72 & 8.57 \\
			& & \textbf{Ours (Pipeline + QA)} & 8.32 & 8.86 & 8.59 \\
			\midrule
			\multirow{3}{*}{\textbf{Qwen3-VL~\cite{yang2025qwen3}}} & \multirow{3}{*}{\textbf{32B}}
			& Direct Generation & 8.18 & 8.65 & 8.41 \\
			& & Full Manual Extension & \best{8.52} & 8.82 & 8.67 \\
			\rowcolor{green!8} 
			& & \textbf{Ours (Pipeline + QA)} & 8.43 & \best{8.94} & \best{8.68} \\
			\bottomrule
		\end{tabular}%
	}
\end{table}

\subsection{In-depth Analysis}

\textbf{Interaction Deduction and Permutation Test.} To evaluate the emergent interaction deduction paradigm (Section~\ref{sec:paradigm_shift}), we compare zero-shot deduction against explicit fusion under Ground Truth (GT) and Predicted (Pred) settings. As shown in Table~\ref{tab:interaction_deduction}, zero-shot text deduction improves upon the reproduced cascaded methods. Recent MOT baselines are listed for completeness, while their interaction scores are not reported because their original outputs do not include the dense descriptions required by this protocol. Notably, our 4B model (w/ Pred Text) achieves a competitive F1 score of 0.515 without task-specific training. To validate the \textit{Macro-Understanding-First} hypothesis, we introduce a prompt permutation test: reversing the token order (\textit{Macro-Last}: instances before environment) or processing them in parallel forces the model to deduce interactions without a global prior, which degrades the F1 score (from 0.515 to 0.472) and reduces Video CIDEr. This result indicates that anchoring the macro-context before micro-dynamics benefits the causal token ordering and supports more accurate logical deduction.

\begin{table}[htbp]
	\centering
	\vspace{-0.1cm}
	\caption{Performance comparison on interaction recognition. A prompt permutation test is included to validate the Macro-First hypothesis.}
	\label{tab:interaction_deduction}
	\resizebox{0.95\linewidth}{!}{%
		\begin{tabular}{l|c|ccc}
			\toprule
			\textbf{Reasoning Paradigm} & \textbf{Model / Information Source} & \textbf{Precision} $\uparrow$ & \textbf{Recall} $\uparrow$ & \textbf{F1} $\uparrow$ \\
			\midrule
			\multirow{4}{*}{\textit{Previous Methods}} 
			& DeepSORT \cite{wojke2017simple} + SMOTer Captioner & 0.365 & 0.277 & 0.310 \\
			& ByteTrack \cite{zhangByteTrackMultiobjectTracking2022} + SMOTer Captioner & 0.443 & 0.258 & 0.326 \\
			& TransTrack \cite{sun2020transtrack} + SMOTer Captioner & 0.406 & 0.311 & 0.352 \\
			& SMOTer \cite{li2024beyond} & 0.434 & 0.320 & 0.368 \\
			\midrule
			\multirow{2}{*}{\textit{Baseline: Explicit Feature Fusion}} 
			& LLMTrack-0.5B + MLP & 0.516 & 0.476 & 0.496 \\
			& LLMTrack-4B + MLP & \best{0.543} & \second{0.516} & \second{0.526} \\
			\midrule
			\textbf{Text Deduction (w/ Pred Text)} & LLaVA-OneVision-0.5B / Generated & 0.492 & 0.468 & 0.480 \\
			\rowcolor{green!10} 
			\textbf{Macro-First (Ours, w/ Pred)} & \textbf{LLaVA-1.5-4B / Video $\rightarrow$ Instance} & 0.528 & 0.502 & \textbf{0.515} \\
			~~ \textit{- Macro-Last (Ablation)} & LLaVA-1.5-4B / Instance $\rightarrow$ Video & 0.482 & 0.462 & 0.472 \\
			~~ \textit{- Parallel (Ablation)} & LLaVA-1.5-4B / Independent Processing & 0.495 & 0.460 & 0.476 \\
			\midrule
			\textbf{Text Deduction (w/ GT Text)} & LLaVA-OneVision-0.5B / GT & 0.522 & 0.495 & 0.508 \\
			\textbf{Text Deduction (w/ GT Text)} & LLaVA-OneVision-1.5-4B / GT & \second{0.536} & \best{0.548} & \best{0.542} \\
			\bottomrule
		\end{tabular}%
	}
	\vspace{-0.4cm}
\end{table}

\textbf{Decoupling Detection from Association.} To verify that our gains stem from superior tracking association rather than a stronger baseline detector, we conduct a controlled experiment (Table~\ref{tab:controlled_association}). Supplying identical Grounding DINO bounding boxes to all baselines normalizes visual perception (DetA $\approx$ 73.1\%) to isolate association capabilities. We additionally include recent association-oriented methods as pending controlled reproductions, since their original detector settings are not directly interchangeable with Grounding DINO. Crucially, traditional methods relying on appearance features use isolated pre-trained ReID networks that lack sequential semantic alignment. In contrast, LLMTrack inherits the query extraction paradigm of FastTrackTr~\cite{liao2025fasttracktr}, deriving instance features directly from tracking queries to deeply unify geometric and semantic representations without secondary extraction. Consequently, under this controlled setting, LLMTrack outperforms the reproduced comparators reported in Table~\ref{tab:controlled_association}, suggesting that transforming inherited query features into continuous semantic trajectory tokens can improve temporal association robustness relative to pure kinematic methods or isolated ReID networks.

\begin{table}[htbp]
    \centering
    \caption{\textbf{Controlled Experiment on Association Capabilities.} To isolate association, all reproduced baselines are provided with the exact GD detections used by LLMTrack. Furthermore, baseline ReID relies on official pre-trained networks, while LLMTrack inherits its query-derived instance features from the paradigm established in FastTrackTr~\cite{liao2025fasttracktr}.}
    \label{tab:controlled_association}
    \resizebox{1.0\linewidth}{!}{%
        \begin{tabular}{l|c|c|cccc}
            \toprule
            \textbf{Method} & \textbf{Detector} & \textbf{Appearance / ReID} & \textbf{HOTA} $\uparrow$ & \textbf{DetA} $\uparrow$ & \textbf{AssA} $\uparrow$ & \textbf{IDF1} $\uparrow$ \\
            \midrule
            SORT \cite{bewleySimpleOnlineRealtime2016} & GD (Ours) & None (Motion + IoU) & 54.2 & 73.1 & 40.2 & 56.8 \\
            DeepSORT \cite{wojke2017simple} & GD (Ours) & Official Pre-trained & 58.4 & 73.1 & 46.6 & 62.2 \\
            ByteTrack \cite{zhangByteTrackMultiobjectTracking2022} & GD (Ours) & None (Motion + IoU) & 71.5 & 73.1 & 69.8 & 79.1 \\
            OC-SORT \cite{cao2023observation} & GD (Ours) & None (Motion + IoU) & 73.2 & 73.1 & 73.3 & 79.5 \\
            AED \cite{fang2025associate} & GD (Ours) & Official Pre-trained & 72.2 & 73.1 & 71.7 & 78.5 \\
            Hybrid-SORT \cite{yang2024hybrid} & GD (Ours) & Official Pre-trained & \second{73.8} & \best{73.1} & \second{74.6} & \second{81.2} \\
            \midrule
            \rowcolor{green!8}
            \textbf{LLMTrack (Ours)} & \textbf{GD (Ours)} & \textbf{Query-derived} & \best{74.6} & 73.1 & \best{76.2} & \best{83.5} \\
            \bottomrule
        \end{tabular}%
    }
    \vspace{-0.2cm}
\end{table}

\subsection{Training Efficiency and Effectiveness}

Training MLLMs on long-video tracking benchmarks like TAO is notoriously memory-intensive. Table~\ref{tab:training_cost} compares the resource consumption of our stage-wise strategy against a standard full-sequence training baseline. By employing sparse frame sampling (Stage 1) and TBPTT with gradient caching (Stage 2/3), we maintain a constant and manageable memory footprint.

\begin{table}[h]
	\centering
	\caption{\textbf{Comparison of Training Efficiency.} We report the GPU memory usage and training duration per epoch on the Grand-SMOT dataset. ``Full Sequence BPTT'' refers to training without the proposed sampling and caching strategies (Batch size = 1 and Only a Single NVIDIA 5880ada).}
	\label{tab:training_cost}
	\resizebox{\linewidth}{!}{
		\begin{tabular}{l|c|c|c}
			\toprule
			\textbf{Method} & \textbf{Memory Complexity} & \textbf{Peak Memory} & \textbf{Time / Epoch}  \\
			\midrule
			Full Sequence BPTT & $\mathcal{O}(T)$ & \textit{OOM} & -  \\
			Random Clipping & $\mathcal{O}(L)$ & - & - \\
			\midrule
			\textbf{Ours (Stage 1)} & $\mathbf{\mathcal{O}(N)}$ & \textbf{16.2 GB} & \textbf{8.6h}  \\ 
			\textbf{Ours (Stage 2/3 0.5B)} & $\mathbf{\mathcal{O}(L)}$ & \textbf{25.8 GB} & \textbf{13.6h}  \\ 
			\textbf{Ours (Stage 2/3 4B)} & $\mathbf{\mathcal{O}(L)}$ & \textbf{36.4 GB} & \textbf{20.1h}  \\ 
			\bottomrule
		\end{tabular}
	}
\end{table}

As presented in Table~\ref{tab:efficiency_ablation}, our training strategy achieves an optimal trade-off between computational efficiency and semantic reasoning performance. Compared to the \textit{Full-Sequence BPTT} baseline, which suffers from Out-Of-Memory (OOM) errors, our Stop-Gradient (SG) caching mechanism bounds GPU memory usage to 25.8 GB. Our TBPTT strategy also mitigates cross-clip context disruption by recursively preserving historical context, restoring the Instance Desc. CIDEr score to 0.355.

\begin{table}[t]
	\centering
	\caption{\textbf{Analysis of Efficiency and Long-term Consistency.} Comparing Decoupled TBPTT against baselines on TAO (0.5B model). Caching ensures long-term consistency, while stop-gradient prevents memory overflow. Removing historical cache severely degrades instance descriptions, proving the necessity of cross-clip context.}
	\label{tab:efficiency_ablation}
	\resizebox{1.0\linewidth}{!}{
		\begin{tabular}{l|c|cc|cc}
			\toprule
			\multirow{2}{*}{\textbf{Training Protocol}} & \textbf{Efficiency} & \multicolumn{2}{c|}{\textbf{Video Summary}} & \multicolumn{2}{c}{\textbf{Instance Desc.}} \\
			\cmidrule(lr){2-2} \cmidrule(lr){3-4} \cmidrule(lr){5-6}
			& \textbf{GPU Mem.} & \textbf{BLEU} $\uparrow$ & \textbf{CIDEr} $\uparrow$ & \textbf{BLEU} $\uparrow$ & \textbf{CIDEr} $\uparrow$ \\
			\midrule
			Full-Sequence BPTT (w/o SG) & \textit{OOM} & - & - & - & - \\
			Independent Clips (w/o Cache) & \textbf{25.4 GB} & 0.126 & 0.269 & 0.115 & 0.218 \\
			Free-running (w/o Teacher Forcing) & 25.8 GB & 0.143 & 0.288 & 0.162 & 0.305 \\
			\midrule
			\rowcolor{green!10} \textbf{Ours (Decoupled TBPTT + TF)} & 25.8 GB & \textbf{0.156} & \textbf{0.315} & \textbf{0.182} & \textbf{0.355} \\
			\bottomrule
		\end{tabular}
	}
\end{table}

	\subsection{LLMTrack as a Vision-Language Foundation}

To validate cross-task generalization and explicitly benchmark against LLMDet~\cite{fu2025llmdet}, we adopt its exact setup to construct a mixed-expert LMM. We concatenate frozen SigLIP~\cite{zhai2023siglip} features with LLMTrack's query-derived semantic features. Using LLaVA-OneVision-0.5b as the base, we pre-train a new projector and fine-tune exclusively on LLaVA-1.5~\cite{li2024llava} (665K). Table~\ref{tab:vqa_generalization} reports performance on GQA~\cite{hudson2019gqa}, POPE~\cite{li2023evaluating}, and MME~\cite{fu2026mme}, with baseline results directly reported from \cite{fu2025llmdet}. LLMTrack consistently surpasses static detector-enhanced models (\textit{e.g.}, MM-GDINO, LLMDet). Our continuous track-level temporal dynamics improve high-level reasoning (MME-Cognition) and suppress hallucinations (POPE) better than isolated frame-level bounding boxes, proving its potential as a robust general multi-modal foundation.
	
\begin{table}[htbp]
	\centering
	\caption{\textbf{Cross-task generalization.} Based on LLaVA-OV-0.5b.}
	\vspace{-0.15cm} 
	\label{tab:vqa_generalization}
	\setlength{\tabcolsep}{2.5pt} 
	\scriptsize 
	\begin{tabular}{l|c|ccc|cc}
		\toprule
		\multirow{2}{*}{\textbf{Method}} & \textbf{GQA} & \multicolumn{3}{c|}{\textbf{POPE}} & \multicolumn{2}{c}{\textbf{MME}} \\
		& \cite{hudson2019gqa} & \textbf{rand} & \textbf{pop} & \textbf{adv} & \textbf{perc.} & \textbf{cog.} \\
		\midrule
		OneVision-0.5b & 56.9 & 87.5 & 86.3 & 85.0 & 1238 & 240 \\
		+ MM-GDINO & \second{61.2} & \second{88.9} & \second{88.1} & \second{86.6} & 1207 & 256 \\
		+ LLMDet & \second{61.2} & 88.8 & 88.0 & 86.0 & \best{1297} & \second{264} \\
		\midrule
		\rowcolor{green!8}
		\textbf{+ LLMTrack (Ours)} & \best{61.6} & \best{89.1} & \best{88.4} & \best{87.0} & \second{1293} & \best{271} \\
		\bottomrule
	\end{tabular}%
	\vspace{-0.4cm} 
\end{table}

	\section{Conclusion}
	
This work challenges prevailing closed-set tracking paradigms by elevating Semantic Multi-Object Tracking (SMOT) to an open-ended generative reasoning task. Our fundamental insight is that multi-object interactions do not require isolated classification tags; rather, they naturally emerge as logical deductions when micro-level individual dynamics are deeply coupled with macro-level environmental contexts. To seamlessly deploy this concept in continuous video streams, we propose LLMTrack. By utilizing a Spatio-Temporal Fusion Module to enforce a \textit{Macro-Understanding-First} generation process, our framework unleashes the potential of MLLMs while effectively suppressing temporal hallucinations. Furthermore, to rigorously validate this paradigm shift, we introduce the Grand-SMOT benchmark. This evaluation platform provides the essential dual-stream dense narratives and a decoupled evaluation protocol to independently assess generative semantic reasoning. Ultimately, we hope our methodology and the open-sourced benchmark will catalyze a paradigm shift in the MOT community, moving beyond passive geometric localization toward active, causal scene comprehension.

\bibliographystyle{IEEEtran}
\bibliography{reference}

\appendix
\setcounter{section}{0}
\setcounter{table}{0}
\setcounter{figure}{0}
\renewcommand{\thesection}{S\arabic{section}}
\renewcommand{\thesubsection}{\thesection.\arabic{subsection}}
\renewcommand{\thetable}{S\arabic{table}}
\renewcommand{\thefigure}{S\arabic{figure}}

    \section{Additional Experimental Details}
    \label{sec:appendix_dataset_stats}

    \subsection{Adaptation of other Trackers}
    \label{sec:appendix_baseline_adaptation}
    
    To ensure a rigorously fair evaluation in the semantic understanding tasks, we established unified adaptation pipelines for both traditional TBD methods (\textit{e.g.}, SORT~\cite{bewleySimpleOnlineRealtime2016}, DeepSORT~\cite{wojke2017simple}, OC-SORT~\cite{cao2023observation}, and ByteTrack~\cite{zhangByteTrackMultiobjectTracking2022}) and query-based trackers (\textit{e.g.}, MOTR~\cite{zeng2022motr}, TransTrack~\cite{sun2020transtrack}, MOTIP~\cite{gao2025multiple}, and AED~\cite{fang2025associate}). Since our LLMTrack natively models and outputs continuous spatio-temporal representations via object queries, we must bridge the architectural gap for legacy baselines to interface with the LLM backend.
    
    For TBD methods, which exclusively output discrete bounding box coordinates and identity associations, we implemented an external feature extraction pipeline. Specifically, we utilize a pre-trained Swin-Transformer-Tiny as a frozen visual encoder. Given an input video frame $I \in \mathbb{R}^{H \times W \times 3}$, the encoder processes the entire image to extract a dense hierarchical feature map $F$:
    \begin{equation}
        F = \text{Swin-Tiny}(I) \in \mathbb{R}^{\frac{H}{32} \times \frac{W}{32} \times C}
    \end{equation}
    where $C = 768$ is the channel dimension of the deepest stage. For each tracked object $i$ with a predicted bounding box $B_i = (x_c, y_c, w, h)$, we apply the standard RoIAlign operation to extract a fixed-size local feature map corresponding to this region:
    \begin{equation}
        f_i = \text{RoIAlign}(F, B_i) \in \mathbb{R}^{K \times K \times C}
    \end{equation}
    where $K=7$ represents the spatial resolution of the pooled RoI feature. A global average pooling (GAP) operation followed by a linear projection layer is then applied to flatten and align the feature into a 1D semantic query vector $q_i$:
    \begin{equation}
        q_i = \text{Linear}(\text{GAP}(f_i)) \in \mathbb{R}^{D}
    \end{equation}
    
    In contrast, adapting DETR-based and end-to-end association methods is significantly more straightforward and aligns closely with the design philosophy of our LLMTrack. Since models like MOTR~\cite{zeng2022motr}, TransTrack~\cite{sun2020transtrack}, OVTrack~\cite{li2023ovtrack}, MOTIP~\cite{gao2025multiple}, and AED~\cite{fang2025associate} inherently maintain high-dimensional track queries or instance embeddings $Q \in \mathbb{R}^{N \times C_{\text{detr}}}$ to represent temporal trajectories, these representations already encapsulate rich semantic and positional information. Therefore, we completely bypass the external Swin-Transformer and RoIAlign steps. For an active track query $Q^{(i)}$ corresponding to object $i$, we directly project it to the LLM's hidden dimension $D$:
    \begin{equation}
        q_i = \text{Linear}\left(Q^{(i)}\right) \in \mathbb{R}^{D}
    \end{equation}
    This resulting token $q_i$ serves as the input query for the LLM fusion module, with the dimension $D$ strictly matching our LLM backend.
    
    We candidly acknowledge that, especially for TBD methods, this post-hoc adaptation is relatively coarse. The bounding-box crops inevitably suffer from background noise inclusion or context truncation, and forcing a geometric tracker's output through an independent visual encoder breaks the end-to-end synergy between tracking and understanding. While the query projection is more natural, it still lacks the explicit spatio-temporal optimization tailored for natural language generation present in LLMTrack. Since engineering an optimal bridging mechanism for legacy trackers is beyond the primary scope of this work, our implementation prioritizes simplicity and fairness over heavy optimization. We believe there is ample room for improvement in this area, and we enthusiastically encourage the community to explore more elegant and robust adapters to endow classical MOT trackers with large language model capabilities.
    
    \begin{table}[htbp]
        \centering
        \caption{\textbf{Standard HOTA evaluation metrics on the Grand-SMOT (TAO Split).} Best results are marked in \best{red}, and second best in \second{blue}. Unlike TETA, which severely penalizes open-vocabulary classification errors, HOTA focuses strictly on geometric localization and association quality.}
        \label{tab:tao_hota}
        \resizebox{\linewidth}{!}{%
            \begin{tabular}{l|cccc}
                \toprule
                \textbf{Method} & \textbf{HOTA} $\uparrow$ & \textbf{DetA} $\uparrow$ & \textbf{AssA} $\uparrow$ & \textbf{IDF1} $\uparrow$ \\
                \midrule
                SORT \cite{bewleySimpleOnlineRealtime2016} & 32.70 & 30.50 & 35.40 & 42.50 \\
                DeepSORT \cite{wojke2017simple} & 34.90 & 31.20 & 39.50 & 45.60 \\
                OC-SORT \cite{cao2023observation} & 44.50 & 35.50 & 55.80 & 56.50 \\
                ByteTrack \cite{zhangByteTrackMultiobjectTracking2022} & 47.70 & 37.80 & \best{60.20} & 60.50 \\
                OVTR \cite{li2025ovtr} & 46.40 & 37.50 & 57.40 & 58.10 \\
                MOTIP \cite{gao2025multiple} & 46.80 & 38.00 & 58.10 & 59.20 \\
                AED \cite{fang2025associate} & 47.50 & 39.00 & 58.80 & 60.10 \\
                OVTrack \cite{li2023ovtrack} & \best{48.40} & \best{40.50} & \second{59.50} & \best{61.20} \\
                \midrule
                \rowcolor{green!8}
                \textbf{LLMTrack (Ours)} & \second{47.90} & \second{39.20} & 58.50 & \second{60.80} \\
                \bottomrule
            \end{tabular}%
        }
    \end{table}
    
    \subsection{Tracking Performance on TAO Split}
    
    As shown in Table~\ref{tab:tao_hota}, traditional tracking algorithms (such as SORT and DeepSORT) obtain lower HOTA and IDF1 scores than the stronger query-based methods on the TAO split. LLMTrack reaches 47.90 HOTA and 60.80 IDF1, closely following the strongest reported TAO result while retaining the capability to generate semantic narratives.
    
    Benefiting from the open-world detection capabilities of Grounding DINO~\cite{liu2024grounding} and query-derived spatio-temporal representations, LLMTrack achieves competitive tracking stability while enabling complex semantic reasoning.
    
    \begin{table}[htbp]
        \centering
        \caption{\textbf{More Multi-object tracking performance comparison on the BenSMOT test set.} Best results are marked in \best{red}, and second best in \second{blue}. For a fair comparison, our LLMTrack also only uses BenSMOT training.}
        \label{tab:appendix_bensmot_tracking}
        \resizebox{\linewidth}{!}{%
            \begin{tabular}{l|cccc}
                \toprule
                \textbf{Method} & \textbf{HOTA} $\uparrow$ & \textbf{DetA} $\uparrow$ & \textbf{AssA} $\uparrow$ & \textbf{IDF1} $\uparrow$ \\
                \midrule
                SORT \cite{bewleySimpleOnlineRealtime2016} & 48.49 & 60.91 & 38.95 & 53.93 \\
                DeepSORT \cite{wojke2017simple} & 50.12 & 61.45 & 40.23 & 56.76 \\
                ByteTrack \cite{zhangByteTrackMultiobjectTracking2022} & 68.84 & 67.10 & 71.15 & 78.37 \\
                MOTR \cite{zeng2022motr} & 66.10 & 55.14 & 73.12 & 68.97 \\
                TransTrack \cite{sun2020transtrack} & 71.31 & 69.67 & 73.34 & 78.67 \\
                OC-SORT \cite{cao2023observation} & 71.74 & 69.12 & 73.55 & 78.22 \\
                AED \cite{fang2025associate} & 71.85 & 65.82 & 76.05 & 73.95 \\
                SMOTer \cite{li2024beyond} & 71.98 & \second{70.79} & 73.71 & \second{80.65} \\
                MOTIP \cite{gao2025multiple} & \second{73.12} & 70.15 & \second{76.31} & 80.12 \\
                \midrule
                \rowcolor{green!8}
                \textbf{LLMTrack (Ours)} & \best{74.61} & \best{73.10} & \best{76.15} & \best{83.52} \\
                \bottomrule
            \end{tabular}%
        }
    \end{table}
    
    \subsection{Semantic Understanding Performance on BenSMOT}
    
    Table~\ref{tab:appendix_understanding_combined} showcases the core competitiveness of each method in the semantic dimension, which is a key distinction between the SMOT task and traditional MOT. Since traditional methods (such as ByteTrack, MOTR, MOTIP, and AED) inherently lack the ability to generate natural language, we construct a rigorously controlled experiment rather than simply attaching an independent, disjointed image captioner. Specifically, the geometric tracking components of all baseline models are trained utilizing their official default settings to ensure optimal spatial localization. For the subsequent semantic generation phase, we equip all baselines (except SMOTer) with our proposed Spatio-Temporal Fusion Module and the exact same 0.5B LLM backend, trained under identical protocols as LLMTrack-0.5B. This setup strictly isolates the quality of the geometric trajectories, proving that our joint optimization paradigm extracts far superior spatio-temporal features for LLM reasoning.
    
    Furthermore, it is worth noting that we report the Semantic Score (GPT-S) exclusively on our Extended BenSMOT split. This independent experiment is trained only on BenSMOT, so its values are not directly comparable to main-paper Table~IV or Appendix Table~S4, which use joint BenSMOT+TAO training. We deliberately omit GPT-S for the Original BenSMOT split because its ground-truth annotations are extremely brief (typically under 20 words). Such rudimentary, template-like short texts are adequately and accurately evaluated by traditional n-gram metrics (\textit{e.g.}, BLEU, CIDEr). Deploying an advanced MLLM judge for these short sentences lacks discriminative value, whereas GPT-S is indispensable for evaluating the logical coherence and hallucination rate of the dense, 100-word narratives in our extended benchmark.
    
\begin{table*}[htbp]
        \centering
        \caption{\textbf{Semantic understanding performance comparison on the BenSMOT test set.} We report results on both the Original (short text) and our Extended (dense narratives) splits. To ensure a strictly fair and rigorous evaluation, \textbf{all baseline tracking methods (except SMOTer) are equipped with our proposed Spatio-Temporal Fusion Module and the exact same 0.5B LLM backend.} Specifically, for TBD baselines marked with an asterisk (*), we extract query features from bounding-box crops using a Swin-Transformer-Tiny. For query/embedding-based trackers marked with a dagger ($\dagger$), we directly utilize their native object queries or instance embeddings for LLM fusion. Best results are marked in \best{red}, and second best in \second{blue}. Note: GPT-S is omitted for the Original split as traditional n-gram metrics are sufficient for evaluating its brief text annotations.}
        \label{tab:appendix_understanding_combined}
        \setlength{\tabcolsep}{3.5pt} 
        \resizebox{\textwidth}{!}{%
            \begin{tabular}{l|ccc|ccc|cccc|cccc}
                \toprule
                \multirow{3}{*}{\textbf{Method}} & \multicolumn{6}{c|}{\textbf{Original BenSMOT (Short Text)}} & \multicolumn{8}{c}{\textbf{Our Extended BenSMOT Split (Dense Narratives)}} \\
                \cmidrule(lr){2-7} \cmidrule(lr){8-15}
                & \multicolumn{3}{c|}{\textit{Video Summary}} & \multicolumn{3}{c|}{\textit{Instance Desc.}} & \multicolumn{4}{c|}{\textit{Video Summary}} & \multicolumn{4}{c}{\textit{Instance Desc.}} \\
                \cmidrule(lr){2-4} \cmidrule(lr){5-7} \cmidrule(lr){8-11} \cmidrule(lr){12-15}
                & \textbf{B-4} & \textbf{M} & \textbf{C} & \textbf{B-4} & \textbf{M} & \textbf{C} & \textbf{B-4} & \textbf{M} & \textbf{C} & \textbf{GPT-S} & \textbf{B-4} & \textbf{M} & \textbf{C} & \textbf{GPT-S} \\
                \midrule
                SORT* \cite{bewleySimpleOnlineRealtime2016} & 0.225 & 0.215 & 0.305 & 0.295 & 0.215 & 0.255 & 0.092 & 0.125 & 0.205 & 2.68 & 0.112 & 0.138 & 0.225 & 1.43 \\
                DeepSORT* \cite{wojke2017simple} & 0.235 & 0.225 & 0.315 & 0.305 & 0.225 & 0.265 & 0.098 & 0.132 & 0.215 & 2.78 & 0.120 & 0.145 & 0.240 & 1.53 \\
                OC-SORT* \cite{cao2023observation} & 0.255 & 0.240 & 0.330 & 0.320 & 0.235 & 0.280 & 0.108 & 0.140 & 0.235 & 3.13 & 0.130 & 0.155 & 0.260 & 1.83 \\
                MOTR$^\dagger$ \cite{zeng2022motr} & 0.280 & 0.255 & 0.360 & 0.345 & 0.245 & 0.300 & 0.120 & 0.150 & 0.255 & - & 0.145 & 0.165 & 0.285 & - \\
                ByteTrack* \cite{zhangByteTrackMultiobjectTracking2022} & 0.295 & 0.260 & 0.370 & 0.365 & 0.250 & 0.315 & 0.125 & 0.155 & 0.265 & 3.00 & 0.150 & 0.170 & 0.295 & 1.70 \\
                TransTrack$^\dagger$ \cite{sun2020transtrack} & 0.315 & 0.275 & 0.395 & 0.385 & 0.265 & 0.335 & 0.138 & 0.165 & 0.285 & - & 0.165 & 0.185 & 0.320 & - \\
                AED$^\dagger$ \cite{fang2025associate} & 0.320 & 0.280 & 0.405 & 0.395 & 0.270 & 0.350 & 0.151 & 0.176 & 0.302 & 3.12 & \second{0.183} & 0.196 & 0.347 & 1.92 \\
                MOTIP$^\dagger$ \cite{gao2025multiple} & 0.325 & 0.285 & 0.410 & 0.405 & 0.275 & 0.355 & 0.155 & 0.180 & 0.310 & 3.22 & 0.181 & 0.202 & 0.352 & 1.94 \\
                SMOTer \cite{li2024beyond} & 0.245 & 0.223 & 0.343 & 0.306 & 0.209 & 0.087 & 0.045 & 0.095 & 0.110 & 2.10 & 0.062 & 0.112 & 0.145 & 1.13 \\
                \midrule
                \rowcolor{green!8}
                \textbf{LLMTrack-0.5B (Ours)} & \second{0.347} & \second{0.302} & \second{0.442} & \second{0.445} & \second{0.296} & \second{0.389} & \second{0.156} & \second{0.185} & \second{0.315} & \second{3.35} & 0.182 & \second{0.205} & \second{0.355} & \second{2.05} \\
                \rowcolor{green!12}
                \textbf{LLMTrack-4B (Ours)} & \best{0.414} & \best{0.381} & \best{0.462} & \best{0.525} & \best{0.342} & \best{0.439} & \best{0.198} & \best{0.224} & \best{0.425} & \best{3.83} & \best{0.235} & \best{0.248} & \best{0.485} & \best{2.95} \\
                \bottomrule
            \end{tabular}%
        }
    \end{table*}
    
    \subsection{Granular Dimension-wise Committee Evaluation}
    
    To comprehensively evaluate the semantic understanding capabilities of our tracking framework and ensure the scientific reproducibility of our metrics, we adopt a multi-dimensional evaluation protocol inspired by the Video-ChatGPT~\cite{maaz2024video} framework. Recognizing that relying on a single proprietary model (e.g., GPT-4o) may introduce subjective bias and risk metric deprecation over time due to API version updates, we upgrade the conventional evaluation to an \textbf{LLM-as-a-Judge Committee}. 
    
    Specifically, we employ an ensemble of four state-of-the-art foundation models to assess the captions generated by our tracking models and baselines: \textbf{GPT-4o}, \textbf{GPT-5.2}, \textbf{Gemini 3.1}~\cite{team2024gemini}, and \textbf{GLM-5}~\cite{zeng2026glm}. To guarantee absolute fairness and deterministic scoring across the committee, we deploy the \textbf{exact same system prompts} and evaluation guidelines for all judge models. Furthermore, we strictly set the API generation \textbf{temperature to 0} to eliminate sampling randomness. This strict protocol ensures a rigorous, reproducible, and objective assessment of complex spatio-temporal reasoning, proving that the superiority of our framework's narratives is acknowledged universally across different LLM architectures.
    
    Table~\ref{tab:appendix_gpt_finegrained} uses the same joint BenSMOT+TAO training setting as main-paper Table~IV. As shown in the table, we break down the overall Semantic Score into five fine-grained dimensions: \textbf{Correctness} (accurate identification of actions and objects), \textbf{Detail Orientation} (richness of attributes and spatial descriptions), \textbf{Contextual Understanding} (logical deduction of scene semantics and social interactions), \textbf{Temporal Understanding} (chronological ordering and causal links of events), and \textbf{Consistency} (absence of logical contradictions or hallucinated trajectories). 
    
    Across both Video and Instance Captioning tasks, \textit{Consistency} generally yields the highest scores across all judges, indicating that most LLM-equipped frameworks can maintain basic narrative logic. Conversely, \textit{Detail Orientation} and \textit{Temporal Understanding} remain severe bottlenecks. Notably, judge models exhibit distinct evaluating characteristics: Gemini 3.1 tends to heavily penalize traditional baselines for temporal fragmentation while rewarding our model's coherent long-term tracking; meanwhile, GLM-5 provides generally more stringent evaluations across all metrics. Despite this variance in scoring strictness, \textbf{LLMTrack-4B} consistently demonstrates a comprehensive superiority across all five dimensions under every evaluating model. This robust consensus among the LLM committee explicitly validates that our natively extracted trajectory representations successfully preserve chronological dynamics.
    
\begin{table*}[htbp]
        \centering
        \caption{\textbf{Fine-grained Multi-Model Semantic Evaluation on Grand-SMOT (BenSMOT Split).} We present the detailed breakdown of the Semantic metrics across five dimensions evaluated by a diverse LLM-as-a-Judge Committee (GPT-4o, GPT-5.2, Gemini 3.1, and GLM-5) with temperature strictly set to 0. The \textbf{Avg.} column is the exact mathematical mean of the five dimensions. The consistent superiority of our method across all judges proves the robustness of the evaluation.}
        \label{tab:appendix_gpt_finegrained}
        \setlength{\tabcolsep}{3.5pt} 
        \resizebox{\textwidth}{!}{%
            \begin{tabular}{l|l|cccccc|cccccc}
                \toprule
                \multirow{3}{*}{\textbf{Method}} & \multirow{3}{*}{\textbf{Judge Model}} & \multicolumn{6}{c|}{\textbf{Video Caption}} & \multicolumn{6}{c}{\textbf{Instance Caption}} \\
                \cmidrule(lr){3-8} \cmidrule(lr){9-14}
                & & \textbf{Corr.} & \textbf{Det.} & \textbf{Ctx.} & \textbf{Temp.} & \textbf{Cons.} & \textbf{Avg.} & \textbf{Corr.} & \textbf{Det.} & \textbf{Ctx.} & \textbf{Temp.} & \textbf{Cons.} & \textbf{Avg.} \\
                \midrule
                
                \multirow{4}{*}{SORT* \cite{bewleySimpleOnlineRealtime2016}} 
                & GPT-4o   & 3.20 & 1.30 & 3.40 & 1.80 & 3.80 & 2.70 & 1.30 & 0.80 & 1.70 & 0.90 & 2.30 & 1.40 \\
                & GPT-5.2  & 3.20 & 1.40 & 3.40 & 1.90 & 3.80 & 2.74 & 1.40 & 0.80 & 1.70 & 1.00 & 2.40 & 1.46 \\
                & Gemini 3.1 & 3.30 & 1.40 & 3.30 & 1.60 & 3.70 & 2.66 & 1.50 & 0.90 & 1.80 & 0.80 & 2.50 & 1.50 \\
                & GLM-5    & 3.10 & 1.20 & 3.30 & 1.70 & 3.70 & 2.60 & 1.20 & 0.70 & 1.60 & 0.80 & 2.20 & 1.30 \\
                \midrule
                
                \multirow{4}{*}{DeepSORT* \cite{wojke2017simple}} 
                & GPT-4o   & 3.30 & 1.40 & 3.50 & 1.90 & 3.90 & 2.80 & 1.40 & 0.90 & 1.80 & 1.00 & 2.40 & 1.50 \\
                & GPT-5.2  & 3.20 & 1.40 & 3.60 & 1.90 & 3.90 & 2.80 & 1.50 & 0.90 & 1.80 & 1.10 & 2.50 & 1.56 \\
                & Gemini 3.1 & 3.40 & 1.50 & 3.50 & 1.70 & 3.80 & 2.78 & 1.60 & 1.10 & 1.90 & 0.90 & 2.60 & 1.62 \\
                & GLM-5    & 3.10 & 1.30 & 3.40 & 1.80 & 3.70 & 2.66 & 1.30 & 0.80 & 1.70 & 0.90 & 2.30 & 1.40 \\
                \midrule
                
                \multirow{4}{*}{ByteTrack* \cite{zhangByteTrackMultiobjectTracking2022}} 
                & GPT-4o   & 3.50 & 1.60 & 3.70 & 2.10 & 4.10 & 3.00 & 1.60 & 1.10 & 2.00 & 1.20 & 2.60 & 1.70 \\
                & GPT-5.2  & 3.50 & 1.70 & 3.80 & 2.20 & 4.10 & 3.06 & 1.70 & 1.10 & 2.00 & 1.30 & 2.60 & 1.74 \\
                & Gemini 3.1 & 3.60 & 1.60 & 3.80 & 2.00 & 4.00 & 3.00 & 1.80 & 1.30 & 2.20 & 1.10 & 2.80 & 1.84 \\
                & GLM-5    & 3.40 & 1.50 & 3.60 & 2.00 & 3.90 & 2.88 & 1.50 & 1.00 & 1.90 & 1.10 & 2.50 & 1.60 \\
                \midrule
                
                \multirow{4}{*}{OC-SORT* \cite{cao2023observation}} 
                & GPT-4o   & 3.60 & 1.70 & 3.80 & 2.30 & 4.10 & 3.10 & 1.70 & 1.20 & 2.10 & 1.30 & 2.70 & 1.80 \\
                & GPT-5.2  & 3.70 & 1.70 & 3.90 & 2.30 & 4.20 & 3.16 & 1.80 & 1.20 & 2.20 & 1.40 & 2.80 & 1.88 \\
                & Gemini 3.1 & 3.80 & 1.80 & 3.90 & 2.20 & 4.20 & 3.18 & 1.90 & 1.40 & 2.30 & 1.20 & 2.90 & 1.94 \\
                & GLM-5    & 3.50 & 1.60 & 3.70 & 2.20 & 4.00 & 3.00 & 1.60 & 1.10 & 2.00 & 1.20 & 2.60 & 1.70 \\
                \midrule
                
                \multirow{4}{*}{Hybrid-SORT* \cite{yang2024hybrid}} 
                & GPT-4o   & 3.60 & 1.80 & 3.80 & 2.20 & 4.10 & 3.10 & 1.80 & 1.20 & 2.10 & 1.30 & 2.60 & 1.80 \\
                & GPT-5.2  & 3.70 & 1.80 & 3.90 & 2.30 & 4.20 & 3.18 & 1.80 & 1.30 & 2.20 & 1.40 & 2.80 & 1.90 \\
                & Gemini 3.1 & 3.80 & 1.80 & 3.90 & 2.30 & 4.20 & 3.20 & 1.90 & 1.40 & 2.30 & 1.20 & 2.90 & 1.94 \\
                & GLM-5    & 3.50 & 1.60 & 3.70 & 2.20 & 4.00 & 3.00 & 1.60 & 1.10 & 2.00 & 1.20 & 2.60 & 1.70 \\
                \midrule
                
                \multirow{4}{*}{OVTR \cite{li2025ovtr}} 
                & GPT-4o   & 3.70 & 1.80 & 3.90 & 2.40 & 4.20 & 3.20 & 1.80 & 1.30 & 2.20 & 1.40 & 2.80 & 1.90 \\
                & GPT-5.2  & 3.80 & 1.80 & 4.00 & 2.50 & 4.20 & 3.26 & 1.90 & 1.30 & 2.30 & 1.50 & 2.80 & 1.96 \\
                & Gemini 3.1 & 3.90 & 1.90 & 4.00 & 2.30 & 4.30 & 3.28 & 2.00 & 1.40 & 2.40 & 1.30 & 3.00 & 2.02 \\
                & GLM-5    & 3.60 & 1.70 & 3.80 & 2.30 & 4.10 & 3.10 & 1.70 & 1.20 & 2.10 & 1.30 & 2.70 & 1.80 \\
                \midrule
                
                \multirow{4}{*}{AED* \cite{fang2025associate}} 
                & GPT-4o   & 3.60 & 1.80 & 3.80 & 2.30 & 4.10 & 3.12 & 1.80 & 1.20 & 2.20 & 1.30 & 2.70 & 1.84 \\
                & GPT-5.2  & 3.70 & 1.80 & 3.90 & 2.40 & 4.20 & 3.20 & 1.80 & 1.30 & 2.30 & 1.40 & 2.80 & 1.92 \\
                & Gemini 3.1 & 3.80 & 1.80 & 3.90 & 2.30 & 4.20 & 3.20 & 1.90 & 1.40 & 2.30 & 1.30 & 2.90 & 1.96 \\
                & GLM-5    & 3.50 & 1.60 & 3.70 & 2.20 & 4.00 & 3.00 & 1.60 & 1.10 & 2.00 & 1.20 & 2.60 & 1.70 \\
                \midrule
                
                \multirow{4}{*}{MOTIP* \cite{gao2025multiple}} 
                & GPT-4o   & 3.80 & 1.90 & 3.80 & 2.30 & 4.10 & 3.18 & 1.70 & 1.40 & 2.30 & 1.30 & 2.70 & 1.88 \\
                & GPT-5.2  & 3.70 & 1.90 & 3.90 & 2.40 & 4.30 & 3.24 & 1.80 & 1.40 & 2.20 & 1.40 & 2.90 & 1.94 \\
                & Gemini 3.1 & 3.80 & 1.80 & 4.10 & 2.40 & 4.20 & 3.26 & 1.90 & 1.50 & 2.30 & 1.40 & 2.90 & 2.00 \\
                & GLM-5    & 3.70 & 1.60 & 3.90 & 2.20 & 4.00 & 3.08 & 1.80 & 1.10 & 2.00 & 1.20 & 2.60 & 1.74 \\
                \midrule
                
                \multirow{4}{*}{OVTrack* \cite{li2023ovtrack}} 
                & GPT-4o   & 3.60 & 1.70 & 3.80 & 2.40 & 4.00 & 3.10 & 1.80 & 1.30 & 2.30 & 1.30 & 2.80 & 1.90 \\
                & GPT-5.2  & 3.70 & 1.80 & 3.80 & 2.50 & 4.10 & 3.18 & 1.80 & 1.40 & 2.30 & 1.40 & 2.90 & 1.96 \\
                & Gemini 3.1 & 3.80 & 1.90 & 3.90 & 2.30 & 4.10 & 3.20 & 2.00 & 1.50 & 2.40 & 1.30 & 3.00 & 2.04 \\
                & GLM-5    & 3.50 & 1.60 & 3.70 & 2.30 & 3.90 & 3.00 & 1.70 & 1.20 & 2.20 & 1.20 & 2.70 & 1.80 \\
                \midrule
                
                \multirow{4}{*}{SMOTer \cite{li2024beyond}} 
                & GPT-4o   & 2.50 & 1.00 & 2.60 & 1.20 & 3.20 & 2.10 & 1.00 & 0.60 & 1.30 & 0.80 & 1.80 & 1.10 \\
                & GPT-5.2  & 2.60 & 1.00 & 2.70 & 1.30 & 3.30 & 2.18 & 1.10 & 0.60 & 1.40 & 0.90 & 1.90 & 1.18 \\
                & Gemini 3.1 & 2.70 & 1.10 & 2.50 & 1.00 & 3.10 & 2.08 & 1.20 & 0.70 & 1.40 & 0.70 & 1.90 & 1.18 \\
                & GLM-5    & 2.40 & 0.90 & 2.50 & 1.10 & 3.10 & 2.00 & 0.90 & 0.50 & 1.20 & 0.70 & 1.70 & 1.00 \\
                \midrule
                
                \rowcolor{green!8}
                \multirow{4}{*}{\cellcolor{white}\textbf{LLMTrack-0.5B (Ours)}} & GPT-4o & 3.80 & 2.00 & 4.10 & 2.50 & 4.10 & 3.30 & 1.90 & 1.40 & 2.40 & 1.50 & 2.80 & 2.00 \\
                \rowcolor{green!8}
                \cellcolor{white} & GPT-5.2 & 3.90 & 2.10 & 4.20 & 2.60 & 4.20 & 3.40 & 2.00 & 1.50 & 2.50 & 1.60 & 2.90 & 2.10 \\
                \rowcolor{green!8}
                \cellcolor{white} & Gemini 3.1 & 4.00 & 2.10 & 4.30 & 2.80 & 4.30 & 3.50 & 2.10 & 1.60 & 2.60 & 1.80 & 3.00 & 2.22 \\
                \rowcolor{green!8}
                \cellcolor{white} & GLM-5 & 3.70 & 1.90 & 4.00 & 2.40 & 4.00 & 3.20 & 1.80 & 1.30 & 2.30 & 1.40 & 2.70 & 1.90 \\
                \midrule
                
                \rowcolor{green!12}
                \multirow{4}{*}{\cellcolor{white}\textbf{LLMTrack-4B (Ours)}} & GPT-4o & 4.40 & 2.60 & 4.70 & 3.00 & 4.30 & 3.80 & 2.80 & 2.20 & 3.30 & 2.30 & 3.90 & 2.90 \\
                \rowcolor{green!12}
                \cellcolor{white} & GPT-5.2 & 4.50 & 2.70 & 4.70 & 3.10 & 4.30 & 3.86 & 2.90 & 2.30 & 3.40 & 2.40 & 4.00 & 3.00 \\
                \rowcolor{green!12}
                \cellcolor{white} & Gemini 3.1 & 4.60 & 2.80 & 4.80 & 3.40 & 4.50 & 4.02 & 3.00 & 2.40 & 3.50 & 2.60 & 4.10 & 3.12 \\
                \rowcolor{green!12}
                \cellcolor{white} & GLM-5 & 4.20 & 2.50 & 4.50 & 2.90 & 4.10 & 3.64 & 2.70 & 2.10 & 3.20 & 2.20 & 3.80 & 2.80 \\
                
                \bottomrule
            \end{tabular}%
        }
    \end{table*}

    \subsection{Qualitative Results}
    \label{sec:appendix_qualitative}
    
    To further demonstrate the open-ended cognitive reasoning and robust tracking capabilities of our framework, we provide qualitative visualizations of LLMTrack on both the BenSMOT and TAO splits.
    
\begin{figure*}[!t]
    \centering
    \includegraphics[width=0.495\textwidth]{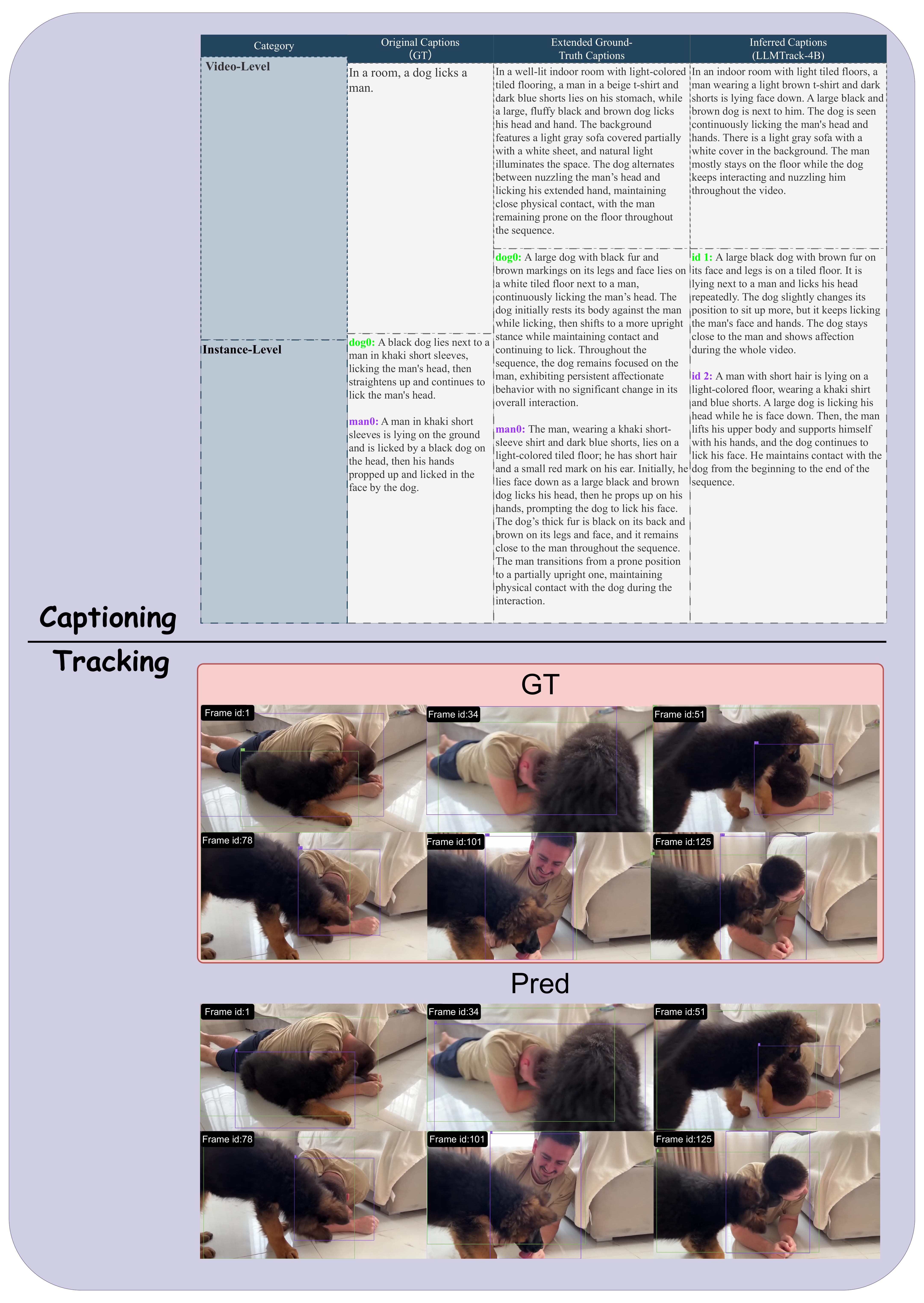}\hfill
    \includegraphics[width=0.495\textwidth]{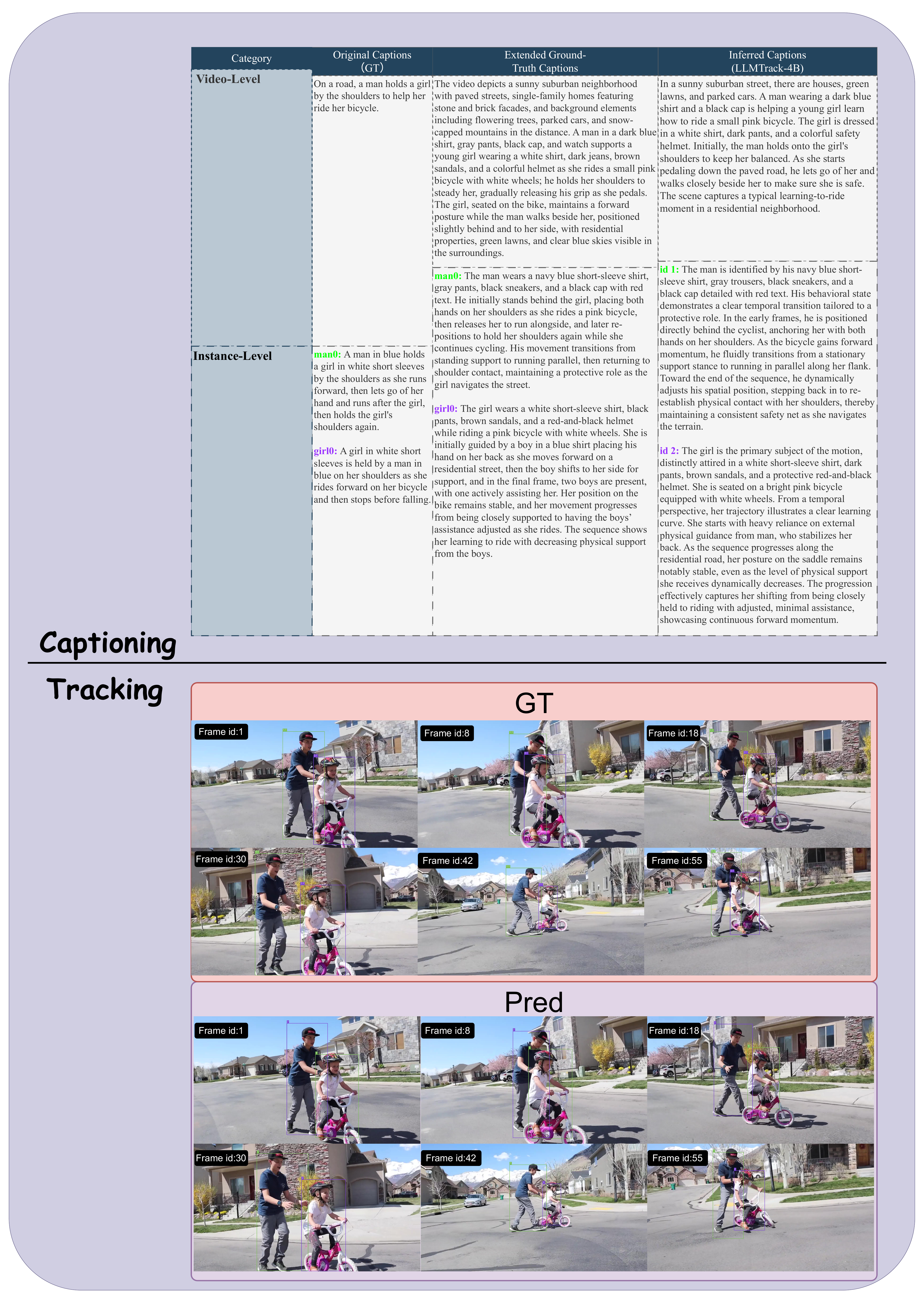}
    \caption{Qualitative semantic tracking results on the BenSMOT dataset.}
    \label{fig:vis_bensmot}
\end{figure*}

\begin{figure*}[!t]
    \centering
    \includegraphics[width=0.40\textwidth]{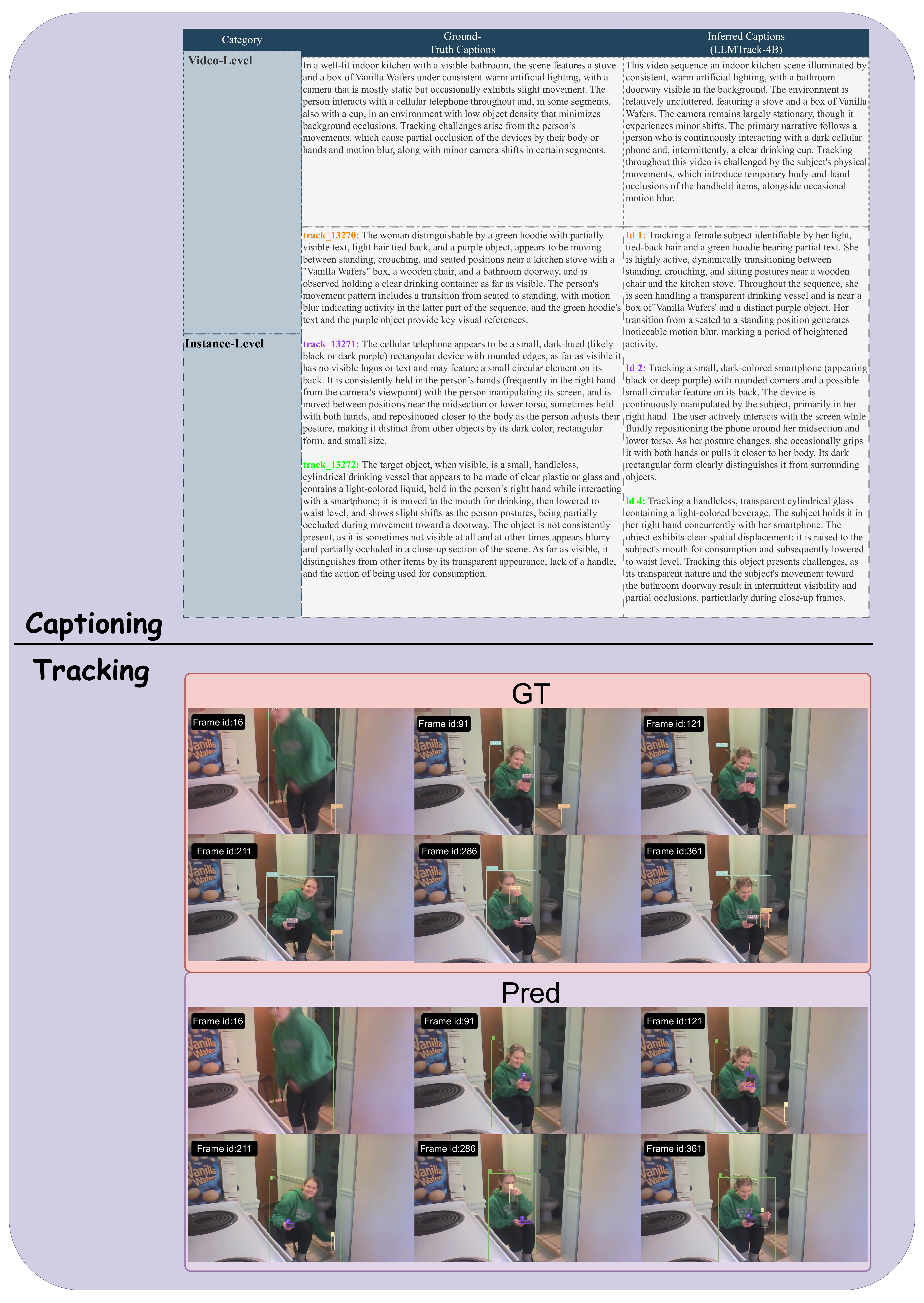}
    \caption{Qualitative semantic tracking results on the TAO split.}
    \label{fig:vis_tao}
\end{figure*}

\end{document}